\documentclass{article}

\usepackage{microtype}
\usepackage{graphicx}
\usepackage{subfigure}
\usepackage{booktabs} 
\usepackage{stfloats} 
\usepackage{placeins} 
\usepackage{paralist}
\usepackage{arydshln}
\usepackage{tikz}

\usepackage[hyphens]{url}
\usepackage[hidelinks]{hyperref}

\usepackage[accepted]{mlsys2026}

\mlsystitlerunning{ExecuTorch - A Unified PyTorch Solution to Run AI Models On-Device}

\begin{document}

\twocolumn[
\mlsystitle{ExecuTorch - A Unified PyTorch Solution to Run AI Models On-Device}



\mlsyssetsymbol{equal}{*}

\begin{mlsysauthorlist}
\mlsysauthor{Mergen Nachin}{equal,meta}
\mlsysauthor{Digant Desai}{equal,meta}
\mlsysauthor{Sicheng Stephen Jia}{equal,meta}
\mlsysauthor{Chen Lai}{equal,formermeta}
\mlsysauthor{Mengwei Liu}{equal,meta}
\mlsysauthor{Jacob Szwejbka}{equal,meta}
\mlsysauthor{Raziel Alvarez}{formermeta}
\mlsysauthor{RJ Ascani}{meta}
\mlsysauthor{Dave Bort}{meta}
\mlsysauthor{Manuel Candales}{meta}
\mlsysauthor{Andrew Caples}{meta}
\mlsysauthor{Yanan Cao}{meta}
\mlsysauthor{Zhengxu Chen}{meta}
\mlsysauthor{Soumith Chintala}{formermeta}
\mlsysauthor{Gregory Comer}{meta}
\mlsysauthor{Tanvir Islam}{formermeta}
\mlsysauthor{Songhao Jia}{meta}
\mlsysauthor{Tarun Karuturi}{meta}
\mlsysauthor{Jack Khuu}{meta}
\mlsysauthor{Abhinay Kukkadapu}{meta}
\mlsysauthor{Tugsbayasgalan Manlaibaatar}{meta}
\mlsysauthor{Andrew Or}{meta}
\mlsysauthor{Kimish Patel}{meta}
\mlsysauthor{Siddartha Pothapragada}{meta}
\mlsysauthor{Lucy Qiu}{meta}
\mlsysauthor{Supriya Rao}{meta}
\mlsysauthor{Orion Reblitz-Richardson}{formermeta}
\mlsysauthor{Max Ren}{formermeta}
\mlsysauthor{Scott Roy}{meta}
\mlsysauthor{Anthony Shoumikhin}{meta}
\mlsysauthor{Scott Wolchok}{formermeta}
\mlsysauthor{Guang Yang}{meta}
\mlsysauthor{Angela Yi}{meta}
\mlsysauthor{Martin Yuan}{meta}
\mlsysauthor{Hansong Zhang}{meta}
\mlsysauthor{Jack Zhang}{meta}
\mlsysauthor{Jerry Zhang}{meta}
\mlsysauthor{Shunting Zhang}{meta}
\mlsysauthor{C.~Cagatay Bilgin}{meta}
\end{mlsysauthorlist}

\mlsysaffiliation{meta}{Meta}
\mlsysaffiliation{formermeta}{Work done while at Meta}

\mlsyscorrespondingauthor{Mergen Nachin}{mnachin@meta.com}
\mlsyscorrespondingauthor{Digant Desai}{digantdesai@meta.com}
\mlsyscorrespondingauthor{Sicheng Stephen Jia}{ssjia@meta.com}

\mlsyskeywords{Machine Learning, MLSys}

\vskip 0.3in

\begin{abstract}
Local execution of AI on edge devices is important for low latency and offline
operation. However, deploying models on diverse hardware remains fragmented,
often requiring model conversion or complete reimplementation outside the PyTorch
ecosystem where the model was originally authored. We introduce ExecuTorch, a
unified PyTorch-native deployment framework for edge AI. ExecuTorch enables
seamless deployment of machine learning models across heterogeneous compute
environments. It scales from embedded microcontrollers to complex
system-on-chips (SoCs) with dedicated accelerators, powering devices ranging
from wearables and smartphones to large compute clusters. ExecuTorch preserves
PyTorch semantics while allowing customization, support for optimizations like
quantization, and pluggable execution ``backends''. These features together
enable fast experimentation, allowing researchers to validate deployment
behavior entirely within PyTorch, bridging the gap between research and
production.
\end{abstract}
]

\printAffiliationsAndNotice{\mlsysEqualContribution}

\section{Introduction}
\label{submission}

Local execution of AI on edge devices is critical for countless important
applications that demand low latency or offline operation, from live translation
to autonomous vehicles and patient monitoring \cite{wang2025optimizing,
wang2025deploying, ng2025development, kuo2020privacy, xu2021federated,
sperling2024reducing, nigade2024inference, kang2024rtswap, pons2023utilization}.
Continued advances in model architectures
\cite{lin2024awqactivationawareweightquantization, vasu2023fastvit,
vasu2024mobileclip} and specialized accelerators such as NPUs
\cite{ahsan2025hardware} have made on-device inference increasingly practical.
However, moving from research to production remains fragmented
\cite{wang2025optimizing, wang2025deploying}: although PyTorch powers over 70\%
\cite{linux_foundation_annual_report_2024} of AI research, ML developers must
either leave the PyTorch environment for platform-specific tools or
accept performance and portability trade-offs.

Deploying AI models on edge devices has spawned numerous solutions, yet existing frameworks suffer from key limitations:
\begin{compactitem}
\item Model conversion between disconnected authoring environments with different semantics (e.g., ONNX~\cite{onnxruntime2018}, TensorFlow Lite~\cite{david2021tensorflowlite})
\item Forced reimplementation in framework-specific formats (e.g., llama.cpp~\cite{gerganov2023llama})
\item Tight coupling to specific hardware vendors (e.g., Qualcomm SNPE~\cite{qualcomm_snpe}, Apple CoreML~\cite{apple2017coreml})
\item Prohibitive runtime overhead (PyTorch Mobile)
\end{compactitem}
These factors create friction in the experimentation-deployment loop. A unified workflow is needed that preserves PyTorch semantics from training to production across all devices without sacrificing performance for portability.

\begin{figure}[htbp]
\centering
\includegraphics[width=1.0\linewidth]{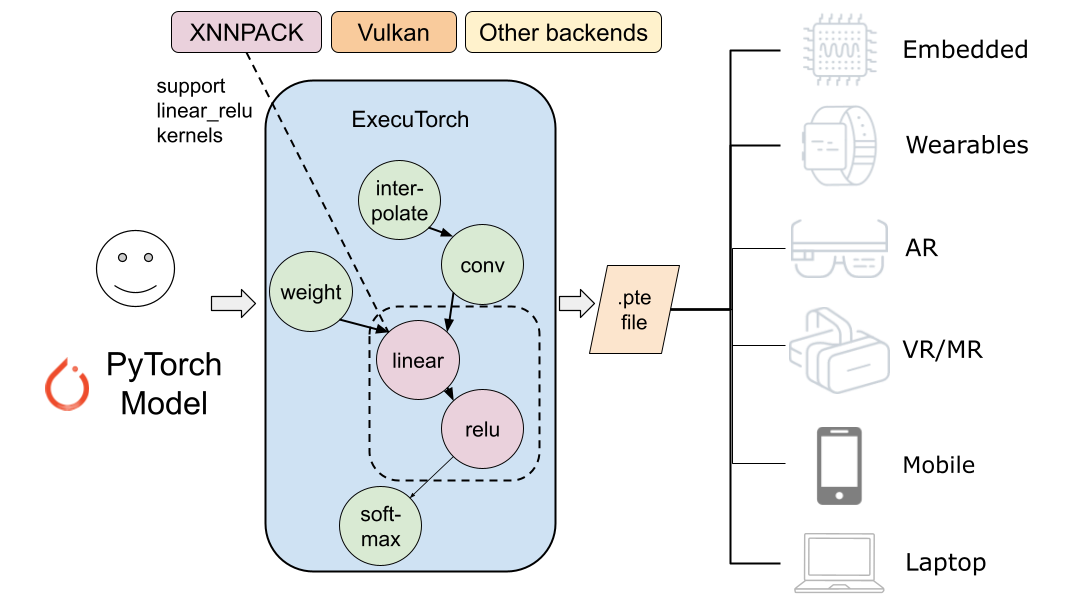}
\caption{Users can bring PyTorch models into ExecuTorch for compilation and optimization (both backend-agnostic and backend-specific) to generate a PTE file that runs on platforms from 0.01 to 800~watts.}
\label{fig:intro}
\end{figure}

ExecuTorch addresses this challenge by providing a PyTorch-native development environment as shown in Figure~\ref{fig:intro}. It leverages PyTorch’s core technology underlying \texttt{torch.compile} and \texttt{torch.export} to construct a portable representation of the original PyTorch model, which can be deployed as a binary across diverse hardware and exploit device-specific kernel optimizations. To do so, ExecuTorch implements infrastructure for \textit{backend delegation}, allowing different parts of the model
to run on the most suitable hardware for a given device. ExecuTorch performs ahead-of-time (AOT) graph-level compilation, which greatly reduces interpreter overhead and runtime dependencies
compared to previous approaches such as TorchScript \cite{pytorch_mobile2019}. As a result, researchers can validate quantization, profile performance, and debug models within PyTorch before deployment.

One advantage of ExecuTorch is experimentation velocity: researchers can validate runtime behavior entirely within PyTorch. Other frameworks' pipelines require model conversion and separate validation, often leading to numerical mismatches and resource-intensive debug cycles. ExecuTorch eliminates this gap by using \texttt{torch.export} to generate execution graphs that can run directly in PyTorch eager mode while capturing a faithful representation that can be executed on-device. For example, a quantized LLM (Large Language Model) can be tested and debugged in PyTorch, then deployed to a phone’s NPU, with confidence that its behavior will match almost exactly.

This parity is achieved through several technologies working in tandem. First, \texttt{torch.export} converts models into hardware-agnostic AOT graphs built from a small set of $<300$ Core ATen primitives, reducing the burden for edge environments. These graphs remove Python dependencies but retain debug symbols, and—unlike ONNX—remain executable in PyTorch for pre-deployment validation. Second, ExecuTorch supports selective backend delegation, allowing parts of a model to run on specialized accelerators like Qualcomm Hexagon or Apple Neural Engine, with CPU fallback as needed. Both AOT and just-in-time compilation modes are supported, and hardware vendors can integrate via a clean API without modifying the core runtime. Third, quantization-aware export makes post-training quantization and quantization-aware training first-class steps. Backends declare their capabilities, and ExecuTorch applies quantization accordingly, ensuring that quantization validated in PyTorch matches on-device execution.

ExecuTorch also addresses memory constraints of on-device LLM deployment through techniques such as KV-cache quantization, sliding-window attention, and 4-bit group-wise weight quantization, which reduces model size by 50\% \cite{metaai2024quantized}. These techniques operate on the model graph produced by \texttt{torch.export}, which allows for evaluation of accuracy and memory trade-offs prior to deployment on a mobile device.

We provide a comprehensive evaluation of ExecuTorch’s latency and throughput across large language models (LLMs) and traditional computer vision models on mobile phones, spanning CPUs, GPUs, and NPUs, and compare it against widely used alternatives such as ONNX Runtime, llama.cpp, and LiteRT. Across devices and workloads, we find that ExecuTorch delivers competitive or state-of-the-art performance across heterogeneous backends: it is consistently strong on CPU (via XNNPACK), achieves high token-generation throughput on mobile GPUs (via Vulkan), and unlocks substantial prefill speedups on NPUs (via QNN) when models are well-delegated, while matching native performance on iOS through CoreML when full-graph delegation is available.

Beyond performance, ExecuTorch provides production-critical customization at two levels: runtime extensions allow developers to implement custom kernels for specialized operations, selectively build operators to reduce binary size, and create custom data loaders for embedded systems; AOT extensions support memory planning and target-specific compiler passes, enabling further optimization for hardware constraints.

This paper makes three principal contributions. First, we present ExecuTorch, the first PyTorch framework to achieve experimentation parity through locally executable export graphs, enabling unified deployment from microcontrollers to smartphones without model conversion or reimplementation. Second, we introduce experimentation parity as a design principle: through \texttt{torch.export} and capability-driven backend integration, researchers validate deployment behavior—quantization, hardware delegation, performance—within PyTorch before committing to production. Third, we demonstrate production viability at scale: ExecuTorch powers billions of daily inferences across Meta’s family of apps \cite{executorch_meta_foa_2025} and Reality Labs (e.g., Ray-Ban smart glasses)\cite{executorch_rl_2025}. It supports execution of vision, audio, and large language models on 12 hardware backends, enabling efficient inference on mobile, embedded, and desktop devices. ExecuTorch demonstrates that researchers need not choose between PyTorch’s development velocity and edge deployment requirements—a unified workflow can deliver both.

\section{Related Work}

Edge AI deployment frameworks make different trade-offs between development
velocity, performance, and portability. We evaluate existing approaches through
the lens of \textit{experimentation parity}, i.e., the ability to validate
deployment behavior within the model development and training environment.

\textbf{Conversion-based frameworks} such as ONNX Runtime \cite{onnxruntime2018}
and TensorFlow Lite \cite{david2021tensorflowlite} decouple training and
deployment through intermediate representations, but the conversion step
introduces semantic gaps that surface only after deployment. For example, QAT in
PyTorch may not translate faithfully to ONNX's quantization semantics. Early
frameworks such as Caffe \cite{jia2014caffeconvolutionalarchitecturefast}
enabled C++-based on-device deployment but required complete model authoring
within their ecosystem.

\textbf{Compiler-based approaches} such as TVM \cite{chen2018tvm} and MNN
\cite{jiang2020mnn} generate optimized kernels through domain-specific
compilation but require learning separate toolchains, tuning procedures, and
debugging workflows distinct from PyTorch.

\textbf{Vendor-specific runtimes}
like Apple's CoreML \cite{apple2017coreml}, Qualcomm's SNPE
\cite{qualcomm_snpe}, and Apple's MLX \cite{mlx2023} deliver excellent
platform-specific performance but fragment the deployment landscape, requiring
parallel implementations for multi-platform support.

\textbf{PyTorch Mobile} \cite{pytorch_mobile2019} and TorchScript
\cite{mobile_interpreter} attempted PyTorch-native deployment but were limited
by high memory footprint and narrow hardware integration.

\textbf{Model-specific runtimes} such as llama.cpp \cite{gerganov2023llama} and
vLLM \cite{kwon2023efficientmemorymanagementlarge} achieve strong performance
through architecture-specific optimization but require complete reimplementation
outside the training framework, breaking iteration velocity. vLLM also requires
a Python runtime, which is not feasible in embedded systems.

\section{Architecture}

ExecuTorch's export APIs and lean runtime allow for seamless deployment of
PyTorch models on any target device, as shown in Figure~\ref{fig:intro}. The key
design goals are to offer:

\begin{compactitem}
\item Unified and Portable Runtime – A lightweight runtime with minimal
dependencies and execution overhead, ensuring maximum portability and
consistent behavior across deployment environments.
\item Composable and Extensible Architecture – Modular interfaces for backends,
graph transform passes, and quantizers allow hardware vendors and developers to
plug in custom components without modifying the core runtime.
\item Efficient Model Execution – Leverage device capabilities and access to
accelerators (CPU, GPU, DSP/NPU), as well as architectural optimizations such as
quantization and memory planning, to minimize memory footprint and latency.
\end{compactitem}

\begin{figure}[htbp]
    \centering
    \includegraphics[width=1.0\linewidth]{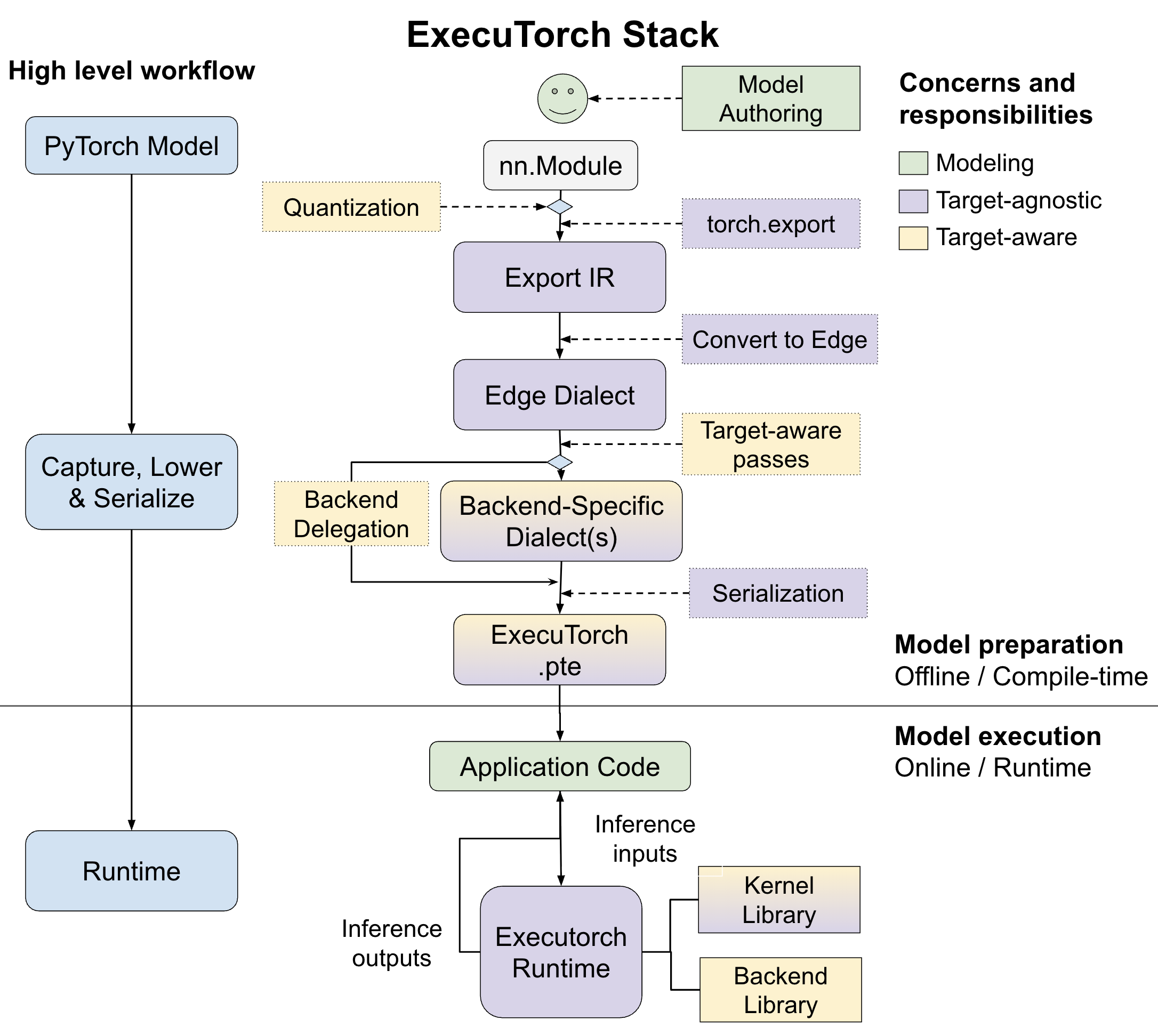}
    \caption{High-level architecture of ExecuTorch, showing two stages: model
    preparation and model execution. The preparation flow exports a PyTorch
    model using \texttt{torch.export}, converts it to the ExecuTorch edge
    dialect, optionally applies backend delegation and graph optimizations, and
    eventually serializes the result into the PTE format for deployment.}
    \label{fig:arch}
\end{figure}

The two main components of ExecuTorch are the AOT export stack and the
Runtime stack (Figure~\ref{fig:arch}).

On the AOT side, ExecuTorch integrates tightly with PyTorch. It uses
\texttt{torch.export} to capture computation graphs from
\texttt{torch.nn.Module} and the \texttt{torch.fx} graph pass infrastructure to
implement graph-level optimizations such as operator fusion. Graph transforms
such as quantization, subgraph delegation, and memory planning are performed
ahead of time, allowing the runtime to stay lean and focus on executing the
pre-optimized model graph.

On the runtime side, ExecuTorch provides a compact and customizable execution
environment. Users can link target-specific kernels and backend libraries to
tailor deployments for specific hardware. The core runtime library is small and
efficient to ensure compatibility with resource-constrained platforms.

\section{Model Preparation}
\label{sec:model-prep}
At a high level, the model preparation workflow follows the steps illustrated by
the diagram in Figure~\ref{fig:arch}.

\subsection{\texttt{torch.export} and IRs}

\texttt{torch.export} is an execution graph capture mechanism provided by
PyTorch. It uses tracing technologies introduced in PyTorch 2.0 \cite{pytorch2}
to convert a model defined in PyTorch Python code into a static graph
data structure, which we call Export IR.

\textbf{Export IR} \cite{torch_export} is a Torch FX graph \cite{torch_fx} with
strong guarantees: (1) Shape soundness: shapes in the captured graph satisfy the shape rules defined by
each operator's semantics; (2) Graph normalization: the graph contains no Python semantics, and
nodes are restricted to a defined operator set; (3) Tensor metadata
availability: shape metadata is available on inputs, intermediate values, and
outputs; (4) Program metadata availability: provenance information records the
original program's \texttt{torch.nn.Module} hierarchy and Python call stack.

Export IR supports progressive lowering into dialects. Subsequent dialects may
enforce custom graph properties and constraints. Complex operators may be
decomposed to enforce a limited operator set; operators may be converted to
functional forms to eliminate mutations and aliasing.

\textbf{Edge Dialect} - ExecuTorch defines a more restrictive Edge Dialect on
top of Export IR with three additional properties: (1)~fully functional graphs
with no mutations or aliasing; (2)~restriction to $<300$ ``Core ATen'' operators
\cite{core_aten_ir}, minimizing the implementation burden for custom kernel
libraries and delegates; and (3)~explicit dtype and memory format
specialization, including a \texttt{dim order} concept that describes the memory
layout of tensors.

Edge Dialect is the IR provided to custom graph passes and delegate lowering
logic. In the final lowering stages, the constraints preventing mutation and
aliasing will be relaxed in very specific scenarios to allow for optimizations
such as KV-cache writeback that require in-place updates. Only non-computational
state updates (i.e., direct data copies without modification) are allowed. Delegates are also allowed to diverge from the
constraints of Edge Dialect during lowering to optimize performance, but they
must ensure that graph transformations produce computations equivalent to the
original Edge Dialect graph.

\subsection{Memory Planning}

Memory planning is performed before serialization as the final preprocessing
step. ExecuTorch analyzes the size and lifespan of each tensor to allocate space
within fixed-size memory arenas, with mutable state tensors given infinite
lifespan to prevent overwriting. The default greedy best-fit algorithm reuses
the smallest non-overlapping buffer when available, but otherwise allocates linearly
to minimize fragmentation. Custom memory planning algorithms are also supported.

\subsection{Quantization}

Quantization is an essential technique for deploying models on device. It significantly
reduces model size and inference latency at some cost to accuracy.
ExecuTorch builds on TorchAO \cite{torchao} to support a variety of backend-specific and generalized quantization algorithms, such as SpinQuant
\cite{liu2025spinquantllmquantizationlearned}. Two quantization workflows are
available: PyTorch 2 Export for static quantization, and Eager mode for dynamic
or weight-only quantization.

\begin{figure}[htbp]
    \centering
    \includegraphics[width=0.8\linewidth]{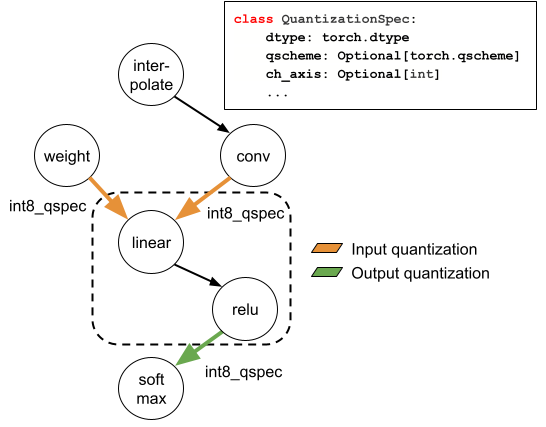}
    \caption{The quantizer annotates input/output tensors of an operator (or pattern) with quantization info such as dtype, bitwidth, range, and observer.}
    \label{fig:pt2e}
\end{figure}

\textbf{PyTorch 2 Export Quantization} \cite{pt2e_ptq, pt2e_qat} transforms a
model in Export IR for both post-training quantization (PTQ) and
quantization-aware training (QAT). The graph is first captured with
\texttt{torch.export}, and each backend uses its own \texttt{Quantizer} class
with annotation APIs to specify quantization intent for operators and patterns,
as shown in Figure~\ref{fig:pt2e}.

\textbf{Eager Mode Quantization} operates directly on \texttt{nn.Module}
instances by converting the weight tensor of target submodules (e.g., linear or
embedding) into a quantized tensor subclass \cite{pytorch_tensor_subclass} or
replacing the submodule with a quantized variant. Each type of quantization
(dtype, packing format) has its own Tensor subclass.

\subsection{Backend Delegate Interface}

\begin{figure}[htbp]
    \centering
    \includegraphics[width=1.0\linewidth]{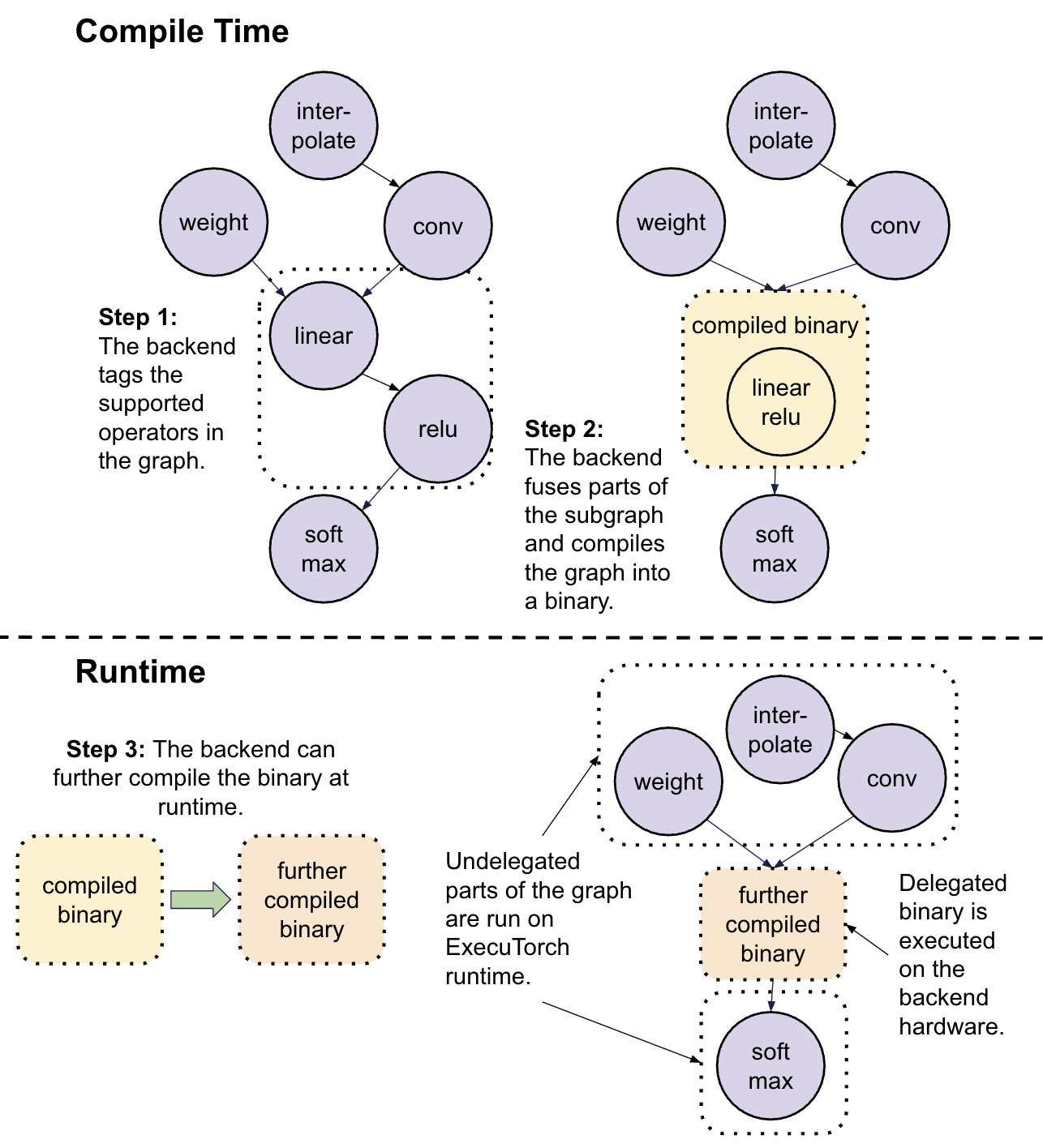}
    \caption{An example showing how the backend receives the graph, compiles it, and executes it.}
    \label{fig:backend_delegate}
\end{figure}

ExecuTorch's backend delegate abstraction enables executing model subgraphs on
specialized processors (Figure~\ref{fig:backend_delegate}). Each delegate
provides (1)~an AOT compiler that lowers compatible Edge Dialect subgraphs into
a backend-specific representation (a ``delegate blob''), and (2)~a runtime library
that can deserialize and execute the delegate blob on the target processor. The
delegate specifies which operators it can accelerate; the partitioner
identifies matching subgraphs composed of supported operators and routes them to
the delegate compiler. This way, backends will not receive subgraphs that they
cannot execute.

\subsection{Model Serialization}
\label{sec:model-prep:model-serialization}

\begin{figure}
    \centering
    \includegraphics[width=0.9\linewidth]{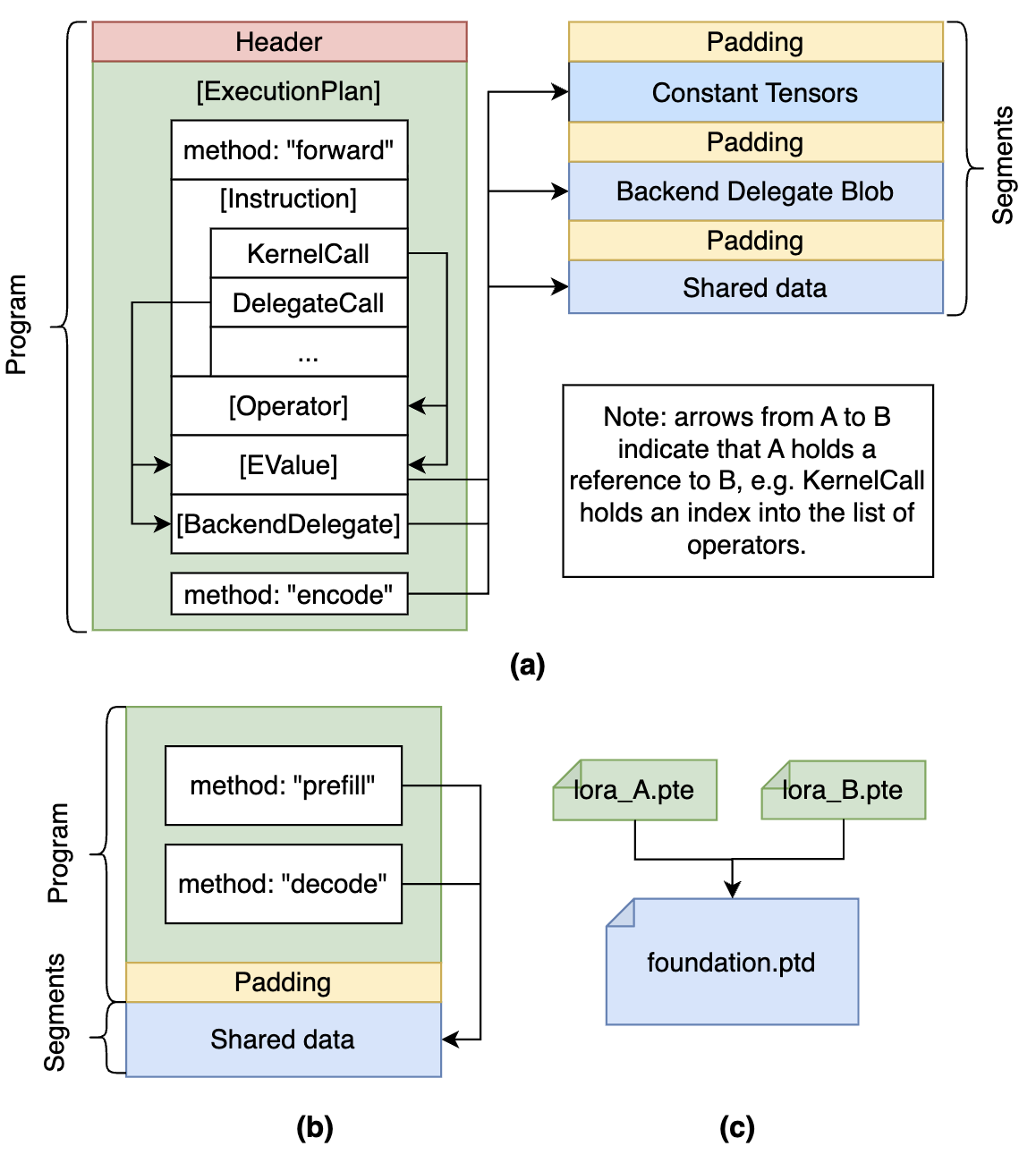}
    \caption{Overview of the PTE file (a) and weight sharing mechanisms; multi-method (b) and program data separation (c).}
    \label{fig:pte-diagram}
\end{figure}

ExecuTorch introduces the PyTorch Edge file format, with the \texttt{.pte} file
extension, designed for minimal runtime overhead and file size
(Figure~\ref{fig:pte-diagram}).

The \textit{program} component contains execution plans for each model method
(e.g., \textit{forward}, \textit{encode}) represented as a list of
instructions. \texttt{KernelCall} invokes an operator, and \texttt{DelegateCall}
invokes execution of a delegated subgraph. Arguments are represented as indices
into a shared list of \texttt{EValue}s, each of which corresponds to a tensor or
scalar. Linear execution (plus the \texttt{Jump} instruction for control flow)
reduces computational overhead compared to executing a graph representation.

The \textit{segments} component contains discrete, aligned memory blocks that
can be independently loaded and freed. Segments holding small program data
persist for the lifetime of the model. Segments representing large delegate
blobs can be freed after model initialization to reduce peak memory.
Page-aligned segments also support direct mmap access without additional
copying.

ExecuTorch also defines the PTD format, with the \texttt{.ptd} file extension,
for storing named tensor and delegate data, enabling weight sharing between PTE
files, checkpointing for on-device training, and independent deployment of
program and data.

\subsection{Weight Sharing}

ExecuTorch supports weight sharing via two mechanisms
(Figure~\ref{fig:pte-diagram}~b,~c): (1)~\textit{multi-method PTE files}, where
different model methods (e.g., LLM prefill and decode) share data segments
within a single file; and (2)~\textit{program-data separation}, where a PTE
program references external PTD weight files that can be shared across models
(e.g., LoRA adapters sharing foundation weights). Both strategies reduce binary
size and enable buffer reuse at runtime.

\subsection{On-Device Fine-Tuning}

ExecuTorch supports on-device training by lowering both forward and backward
execution graphs. Updated weights are written as new PTD checkpoints. We
validated fine-tuning a classification model (lowered to the XNNPACK backend)
using the CIFAR-10 dataset on an Android device.

\section{Model Execution}

\begin{figure}
\centering
\includegraphics[width=0.8\linewidth]{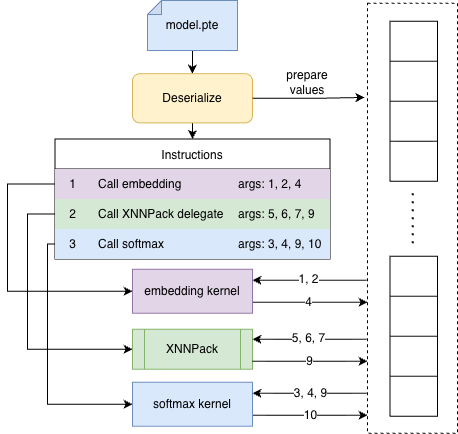}
\caption{ExecuTorch Runtime}
\label{fig:runtime}
\end{figure}

ExecuTorch provides a lightweight and modular runtime with tight memory and
compute budgets (Figure~\ref{fig:runtime}). The runtime executes the instruction
lists encoded in the \texttt{program} component of the PTE file
(Section~\ref{sec:model-prep:model-serialization}), where each instruction maps
to a statically registered kernel or delegate call. At build time, the selective
build API allows developers to link only the required kernel and delegate
libraries. Static registration removes dynamic operator resolution overhead and
minimizes per-instruction latency.

\subsection{Core Runtime Portability}

The core runtime targets C++17 and does not depend on dynamic memory allocation,
synchronization primitives, or exceptions in order to maximize portability across
hardware platforms (servers, mobile phones, and bare-metal embedded systems) and
software environments (POSIX, Windows, bare-metal environments). Note that these
restrictions do not apply to extensions and backends, which may be intended for
use only on specific platforms.

The core runtime does not create or manage heap-allocated memory or use C++ STL
types which allocate or manage their own memory. All memory must be user
provided via the \texttt{MemoryManager} abstraction which, in addition to
ensuring portability, enables placement of tensor data in specialized memory
regions such as SRAM or DRAM. The \texttt{FreeableBuffer} abstraction enables
lifetime management by wrapping a buffer pointer and a user-defined free
function.

The Platform Abstraction Layer (PAL) allows overriding system operations such as
logging, querying the time, or panicking the process/system. The
\texttt{DataLoader} interface supports custom PTE loading strategies (e.g., file
I/O or mmap).

\subsection{Runtime API Language Bindings}

ExecuTorch ships optional extensions that present a PyTorch-like façade over the
core runtime, including a C++ \texttt{Module} API mirroring eager-mode usage and
\texttt{TensorPtr} for safe, zero-copy tensor passing. Native bindings for iOS
(Objective-C/Swift) and Android (Java/Kotlin) allow apps to call into ExecuTorch
without touching C++ directly.

\subsection{Runtime Overhead}

To quantify reduction in framework overhead, we compare inference of a minimal
model (\texttt{mul} + \texttt{add}) between ExecuTorch and the TorchScript-based
PyTorch Mobile Interpreter~\cite{mobile_interpreter}. FlatBuffer-based
deserialization yields a 5.3$\times$ speedup, runtime initialization is
37.4$\times$ faster by eliminating dynamic operator resolution, and simplified
instruction-to-kernel routing reduces per-operator overhead. Deterministic
memory behavior follows from ahead-of-time memory planning.

\begin{table}[ht]
\centering
\small
\caption{Runtime overhead comparison between ExecuTorch and the PyTorch Mobile
Interpreter on a minimal model. All values in CPU cycles.}
\label{tab:runtime_overhead}
\resizebox{\columnwidth}{!}{%
\begin{tabular}{llrrr}
\toprule
\textbf{Phase} & \textbf{Component} & \textbf{MI} & \textbf{ET} &
\textbf{Speedup} \\
\midrule
Loading   & Deserialization      & 510       & 97      & 5.3$\times$    \\
Loading   & Initialization       & 312{,}631 & 8{,}350 & 37.4$\times$   \\
\cmidrule(lr){1-5} Execution & Framework overhead   & 324{,}399 & 75      &
4{,}325$\times$ \\
Execution & \texttt{aten::mul}   & 7{,}976   & 360     & 22.0$\times$   \\
Execution & \texttt{aten::add}   & 8{,}493   & 390     & 21.8$\times$   \\
\cmidrule(lr){1-5} \multicolumn{2}{l}{\textbf{Total per inference}} & 654{,}009
& 9{,}272 & 70.5$\times$ \\
\bottomrule
\end{tabular}%
}
\end{table}

\section{Kernels}

ExecuTorch kernels are stateless functions that implement tensor operations
characterized by a fixed name, an operator schema, and clear input and output
semantics (i.e.,\ expected data types, aliasing, etc.). Built-in kernel libraries
(i.e.,\ a collection of kernels that is linked during build and invoked during
execution) implement the Core ATen operator set used in Edge Dialect. Operator
semantics are identical to the corresponding implementations in PyTorch's ATen
library, except that tensor memory must be densely packed. As in PyTorch, custom
operators may be defined using PyTorch's custom operator API.

\subsection{Kernel Libraries}

ExecuTorch ships with two CPU kernel libraries. The \textit{Portable Kernel
Library} is a reference implementation of Core ATen operators with no external
dependencies. It provides functional and correct implementations that are always
available. The \textit{Optimized Kernel Library} accelerates selected operators
using SIMD intrinsics and optimized math libraries (e.g., SLEEF, OpenBLAS),
trading portability for performance. Users may map operators to either library.

\subsection{Kernel Registration APIs}

\textbf{Selective Build:} The full portable library is $\sim$2.3~MiB, which is too
large for many resource-constrained applications. ExecuTorch's selective build
feature allows users to specify a subset of kernels to include when building,
which can reduce binary size from MiB to KiB. To further reduce binary size,
dtype selective build preserves only kernel code paths for data types that will
actually be exercised during inference, discarding the rest.

\textbf{Runtime Registration API:} In PyTorch, operator schemas are resolved by
parsing a string DSL at static initialization time, incurring significant
startup latency. ExecuTorch avoids this by capturing and storing argument sequence and type
information based on the exported graph. To account for the possibility
of PyTorch operator schema/functionality being updated, Edge Dialect operators
have strong backward compatibility guarantees.

\section{Backends}

ExecuTorch currently includes a diverse selection of production-ready backends,
which together enable efficient execution across a wide range of processors.
Additionally, several more backends are under active development: MediaTek,
OpenVINO, Samsung Exynos, NXP eIQ Neutron, CUDA, and Metal.

\subsection{XNNPACK CPU Backend}

The XNNPACK backend is a high-performance CPU backend built on Google's XNNPACK
library \cite{xnnpack}. Active collaboration between the ExecuTorch and Google
teams ensures tight integration and alignment between the two projects.
Complementary libraries such as Arm's KleidiAI \cite{kleidiai} are also used to
extend hardware coverage. The quantizer exposes support for static and dynamic
quantization for int8 inference, as well as per-channel and group-wise int4
weights for LLM workloads.

At runtime, the delegate achieves high performance using an extensive library of
SIMD-optimized, multithreading-compatible kernels tuned across a wide range
of CPU architectures and input shapes. A weight caching mechanism enables
efficient model reloading and LoRA weight sharing.

\subsection{Vulkan Backend}

The Vulkan \cite{vulkan_spec} backend is designed for inference on mobile GPUs.
Model inference is powered by a growing library of GLSL compute shaders that
currently implement 76 ATen operators. One operator may map to multiple compute
shaders, which are selected at runtime based on tensor storage type, memory
layout, supported Vulkan extensions, input shapes, and GPU architecture. Similar
to XNNPACK, the backend also includes int8 variants of several operators
(convolution, matmul, linear) to support int8 inference via static or dynamic
quantization. Integer inference leverages
hardware-accelerated integer dot product instructions when available. Inference
with group-wise quantized int4 weights is also supported for LLM workloads.

\subsection{Arm Ethos-U NPU Backend}

The Arm backend targets Ethos-U NPUs \cite{arm_ethosu}, including the latest
Ethos-U85, for efficient inference on embedded platforms. The delegate
converts Edge Dialect subgraphs to TOSA (Tensor Operator Set Architecture)
\cite{tosa_spec} IR, which is then compiled by the Vela compiler (Regor backend)
into optimized binaries for Ethos-U NPUs. The backend's quantizer features
symmetric \texttt{int8} and mixed-precision quantization to support a wide range
of models.

\subsection{Qualcomm QNN Backend}

The QNN backend, built on Qualcomm AI Engine Direct
\cite{qualcomm_ai_engine_direct}, targets inference on the Hexagon DSP using the
A8W8 and A16W4 quantization formats. While static shapes offer optimal
performance, limited dynamic shape support is also available. The backend is
compatible with many quantization algorithms and features in TorchAO, including
SpinQuant, Range-Setting, and a shared SeqMSE observer. It supports multi-method
execution, spill-fill buffers, runtime-adjustable power modes, profiling, and both
offline and online compilation.

\subsection{CoreML Backend}

The CoreML backend \cite{coreml} targets inference on Apple platforms. It is
able to perform model inference on CPU, GPU, and Neural Engine (ANE) processors.
It exposes most capabilities available when using CoreML directly, such as 8-bit
static quantization and weight-only quantization; selection of compute units and
compute precision; support for static, enumerated, and dynamic shapes; and
execution of stateful models.

\section{Devtools}

ExecuTorch provides a suite of developer tools for profiling and debugging
on-device deployments. These tools assist developers in linking the eager-mode
behavior of a model in PyTorch to the on-device behavior when executing with
ExecuTorch. The workflow centers on two artifacts: \texttt{ETRecord}, produced
during export, captures the Edge Dialect graph and contains debug metadata linking
runtime events to the Python source code from the original model; and
\texttt{ETDump}, produced during execution, captures operator latencies, memory
lifetimes, delegate and kernel call events, and optionally intermediate tensor
values.

The Inspector API then provides tools to view and analyze the data contained in
these artifacts. Developers can use latency data to identify slow operators and
bottlenecks within delegates. When a model is producing incorrect outputs,
intermediate tensor values can be inspected and compared to a reference model to
identify the source of the error. Tensor lifetimes and allocator behavior can
also be analyzed to reduce memory footprint.

\section{Enabling Use Cases}

\subsection{Large Language Models}

LLMs can be exported to ExecuTorch using either ExecuTorch's modular
transformer-decoder definition, which supports popular LLMs like Llama 3.2 and
Qwen3 \cite{yang2025qwen3technicalreport}, or Hugging Face’s Transformers
library \cite{wolf-etal-2020-transformers} using Optimum ExecuTorch
\cite{HuggingFace_optimum_executorch}. Over 80\% of models on Hugging Face’s text
generation leaderboard can be exported through Optimum ExecuTorch.

To represent a LLM's prefill and decode steps with a single graph, \texttt{torch.export}
marks the sequence length dimension as dynamic. For backends that
require static shapes (e.g., QNN), two separate models are exported: one for
prefill, padded to the maximum context length, and one for single-token decode.
ExecuTorch provides a C++ tokenizer that can be deployed to native platforms
with minimal dependencies.

\begin{figure}[htbp]
    \centering
    \includegraphics[width=1.0\linewidth]{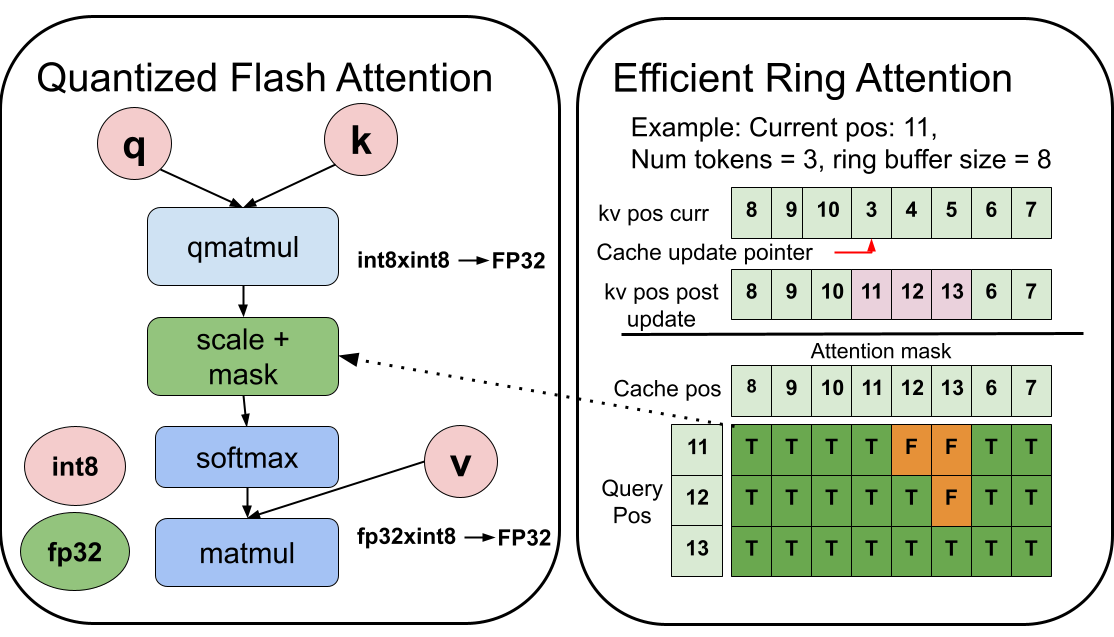}
    \caption{(a) Quantized Flash Attention and (b) Efficient sliding window attention}
    \label{fig:llm_opt}
\end{figure}

Key optimizations accelerate LLM inference on CPU (Figure~\ref{fig:llm_opt}):

\textbf{Flash attention}: Avoids the cost of materializing intermediate
attention tensors, which is particularly useful for reducing memory footprint and
inference latency for long contexts.

\textbf{Quantized KV cache and attention}: A per-channel quantized KV cache with
quantized attention reduces memory for long contexts
(Figure~\ref{fig:llm_opt}~(a)).

\textbf{Efficient sliding window attention}: For models with local-global
attention \cite{shao2024localglobalattentionadaptivemechanism} (e.g., Gemma 3),
cache positions are tracked in a separate array to generate causal masks without
shifting the KV cache (Figure~\ref{fig:llm_opt}~(b)), avoiding costly memory
movement.

The QNN and CoreML backends also support speculative decoding as an optimization
for token generation throughput.

\subsection{Multi-modality}

ExecuTorch supports multi-modal transformers by splitting models at export time
into text embedding, text decoder, and multi-modal encoder components. They are
then efficiently stitched together in the C++ model runner. This targets Early
Fusion models \cite{9190246} (e.g., Voxtral \cite{liu2025voxtral}, Gemma 3 4B
\cite{gemmateam2025gemma3technicalreport}); cross-attention models are supported
via an attention interface for external KV cache management.

\subsection{MCU Deployment}

ExecuTorch's portable runtime and selective build system enable deployment on
microcontroller-class targets. We demonstrate this with an MNIST digit
classifier on the Raspberry Pi Pico 2 (Arm Cortex-M33, 520~KiB SRAM, 4~MiB
Flash), comparing two configurations: FP32 inference using the Portable Kernel
Library, and int8 inference using Arm's CMSIS-NN~\cite{cmsis_nn} library via
the Arm Ethos-U backend.

\textbf{Selective build} reduces runtime code size by linking only the kernels
required by the model. For the FP32 Portable configuration, selective build
shrinks total code size from 1{,}322~KiB to 253~KiB (5.2$\times$); for int8
CMSIS-NN, from 1{,}248~KiB to 203~KiB (6.2$\times$).

Table~\ref{tab:mcu_flash} provides a detailed Flash breakdown with selective
build enabled. The ExecuTorch runtime itself occupies only 13--26~KiB; the
majority of Flash is consumed by the model artifact, kernel libraries, and
system code (Pico SDK + libc).

\begin{table}[ht]
\centering
\small
\caption{Flash breakdown with selective build (KiB).}
\label{tab:mcu_flash}
\begin{tabular}{lrr}
\toprule
\textbf{Component} & \textbf{FP32 Portable} & \textbf{INT8 CMSIS-NN} \\
\midrule
Model (.pte)             & 103.7 & 29.1  \\
ET Runtime               & 25.7  & 13.1  \\
Kernel Registration      & 0.3   & 2.9   \\
CMSIS-NN Library         & ---   & 35.8  \\
Cortex-M Ops             & ---   & 5.9   \\
System (Pico SDK + libc) & 123.3 & 116.0 \\
\midrule
\textbf{Total Flash}     & 253   & 203   \\
\bottomrule
\end{tabular}
\end{table}

Table~\ref{tab:mcu_ram} reports measured RAM usage. Ahead-of-time memory
planning determines the arena size at export time, eliminating runtime
allocation. Int8 quantization and operator fusion reduce the memory-planned
arena from 101.2~KiB to 3.8~KiB, bringing total RAM to approximately 11~KiB---
well within the Pico~2's 520~KiB SRAM budget.

\begin{table}[ht]
\centering
\small
\caption{RAM breakdown, measured on device (KiB).}
\label{tab:mcu_ram}
\begin{tabular}{lrr}
\toprule
\textbf{Component} & \textbf{FP32 Portable} & \textbf{INT8 CMSIS-NN} \\
\midrule
Memory Planned Arena & 101.2     & 3.8       \\
Method Allocator     & $\sim$2--3 & $\sim$2--3 \\
Kernel Registry      & 0.2       & 0.2       \\
Other BSS            & 4.2       & 4.2       \\
\midrule
\textbf{Total RAM}   & $\sim$108 & $\sim$11  \\
\bottomrule
\end{tabular}
\end{table}

Int8 quantization with CMSIS-NN delivers a 16.46$\times$ inference speedup
(3.5~ms vs 57.6~ms) while reducing model size by 3.6$\times$ and RAM by
10$\times$ compared to the FP32 configuration. These results demonstrate that
ExecuTorch can deploy real models on sub-dollar MCUs.

\section{Platform and Hardware Support}

ExecuTorch supports diverse platforms via source-level portability and multiple
hardware backends. The core runtime builds on Windows and Unix using Clang or
GCC. Embedded targets use standard C++ with minimal toolchain
assumptions and macro-based abstractions for compiler extensions. Platform
bindings for Android and iOS offer out-of-the-box usability, while backends may
relax portability to leverage hardware-specific optimizations.

\begin{table}[ht]
\centering
\small
\caption{Platform compatibility.}
\label{tab:platform_compat}
\resizebox{\columnwidth}{!}{%
\begin{tabular}{llllll}
\toprule
\textbf{Platform} & \textbf{Backends} & \textbf{Language} \\
\midrule
Windows         & Vulkan, CUDA, XNNPACK        & C++           \\
Linux           & Intel OpenVINO                               \\
\midrule
MacOS           & CoreML, MPS, XNNPACK        & C++, Swift,   \\
iOS             &                              & Objective-C   \\
\midrule
Android         & Vulkan, XNNPACK, Arm VGF,    & C++, Java,    \\
                & Qualcomm NPU, MediaTek NPU,  & Kotlin        \\
                & Samsung Exynos               &               \\
\midrule
Embedded Systems& Cortex-M, Arm Ethos-U,       & C++           \\
                & NXP NPU, Cadence DSP                         \\
\bottomrule
\end{tabular}%
}
\end{table}

Table~\ref{tab:platform_compat} summarizes platform and backend coverage. For
platforms without an accelerated backend, the portable kernel library serves as
a fallback.

\section{Performance Evaluations}

We evaluate ExecuTorch's performance on a representative set of large language
models and image classification models. We compare against other widely adopted
on-device ML frameworks: llama.cpp, ONNX Runtime, LiteRT, and CoreML. The most
recent framework versions as of March 31, 2026, are used. Experiments were
conducted on a Samsung Galaxy S25 Ultra (Snapdragon 8 Elite SoC; 16 GiB RAM, 2×
Cortex-X925 + 6× Cortex-A725 CPUs, Adreno 830 GPU, Hexagon NPU), a Google Pixel
9 Pro XL (Tensor G4 SoC; 16 GiB RAM, 1× Cortex-X4 + 3× Cortex-A720 + 4×
Cortex-A520 CPUs, Mali-G715 GPU, Edge TPU), and an Apple iPhone 15 Pro. Missing
entries (``–'') indicate configurations for which data could not be collected due
to issues during model export or inference.

\begin{table*}[t]
\centering
\small
\caption{ExecuTorch (ET) prefill and decode throughput range in tokens/sec, and model size (i.e., the ``Size'' column) in MiB for Qwen3 0.6B, Llama 3.2 1B, and Phi4 Mini compared against other frameworks. The ``GS'' column indicates the quantization group size.}
\label{tab:llm_tps}
\resizebox{\textwidth}{!}{%
\begin{tabular}{llc rr rr r rr rr r rr rr r}
\toprule
& & & \multicolumn{5}{c}{\textbf{Qwen3 0.6B}} & \multicolumn{5}{c}{\textbf{Llama 3.2 1B}} & \multicolumn{5}{c}{\textbf{Phi4 Mini (3.8B)}} \\
\cmidrule(lr){4-8} \cmidrule(lr){9-13} \cmidrule(lr){14-18}
& & & \multicolumn{2}{c}{\textbf{Prefill}} & \multicolumn{2}{c}{\textbf{Decode}} & & \multicolumn{2}{c}{\textbf{Prefill}} & \multicolumn{2}{c}{\textbf{Decode}} & & \multicolumn{2}{c}{\textbf{Prefill}} & \multicolumn{2}{c}{\textbf{Decode}} & \\
\cmidrule(lr){4-5} \cmidrule(lr){6-7} \cmidrule(lr){9-10} \cmidrule(lr){11-12} \cmidrule(lr){14-15} \cmidrule(lr){16-17}
\textbf{Hardware} & \textbf{GS} & \textbf{Framework} & \textbf{min} & \textbf{max} & \textbf{min} & \textbf{max} & \textbf{Size} & \textbf{min} & \textbf{max} & \textbf{min} & \textbf{max} & \textbf{Size} & \textbf{min} & \textbf{max} & \textbf{min} & \textbf{max} & \textbf{Size} \\
\midrule
\multicolumn{18}{l}{\textit{Samsung Galaxy S25 Ultra (Snapdragon 8 Elite)}} \\
\midrule
CPU & 32 & ET XNNPACK & 716.62 & \textbf{732.59} & 72.34 & \textbf{72.73} & 417 & 524.59 & \textbf{528.93} & 65.88 & \textbf{67.10} & 821 & 143.60 & \textbf{159.70} & 18.50 & \textbf{19.90} & 2428 \\
& & llama.cpp & 747.70 & \textbf{750.40} & 97.90 & \textbf{100.80} & 442 & 512.70 & \textbf{537.80} & 65.60 & \textbf{66.50} & 728 & 151.20 & \textbf{153.30} & 22.10 & \textbf{22.90} & 2216 \\
& & ONNX & 387.68 & \textbf{443.93} & 60.32 & \textbf{65.09} & 376 & 284.12 & \textbf{328.76} & 60.37 & \textbf{64.73} & 769 & 60.02 & \textbf{65.72} & 17.37 & \textbf{18.29} & 2337 \\
& & LiteRT & 150.83 & \textbf{153.38} & 11.88 & \textbf{12.48} & 326 & 143.49 & \textbf{149.63} & 31.36 & \textbf{33.03} & 667 & - & - & - & - & - \\
\noalign{\vskip 3pt}
\hdashline
\noalign{\vskip 3pt}
& 128 & ET XNNPACK & 837.58 & \textbf{848.39} & 74.92 & \textbf{75.25} & 376 & 649.75 & \textbf{658.10} & 71.69 & \textbf{71.97} & 742 & 195.70 & \textbf{202.60} & 20.50 & \textbf{22.00} & 2201 \\
& & ONNX & 483.27 & \textbf{554.74} & 63.52 & \textbf{65.39} & 323 & 490.63 & \textbf{538.20} & 74.60 & \textbf{76.56} & 659 & 92.08 & \textbf{106.54} & 18.81 & \textbf{20.02} & 1994 \\
& & LiteRT & 172.57 & \textbf{173.22} & 12.56 & \textbf{13.23} & 299 & 185.35 & \textbf{190.26} & 34.30 & \textbf{34.56} & 612 & - & - & - & - & - \\
\cmidrule(lr){1-18}
GPU & 32 & ET Vulkan & 1206.42 & \textbf{1246.45} & 57.98 & \textbf{58.23} & 457 & 927.54 & \textbf{930.91} & 59.19 & \textbf{59.40} & 920 & 191.20 & \textbf{238.92} & 16.03 & \textbf{16.38} & 2829 \\
& & llama.cpp & 1709.90 & \textbf{1718.00} & 77.80 & \textbf{78.80} & 442 & 1064.80 & \textbf{1092.70} & 36.70 & \textbf{42.00} & 728 & 339.10 & \textbf{341.50} & 11.60 & \textbf{13.10} & 2216 \\
& & ONNX & - & - & - & - & - & - & - & - & - & - & - & - & - & - & - \\
& & LiteRT & - & - & - & - & - & - & - & - & - & - & - & - & - & - & - \\
\noalign{\vskip 3pt}
\hdashline
\noalign{\vskip 3pt}
& 128 & ET Vulkan & 1538.01 & \textbf{1556.21} & 62.35 & \textbf{63.52} & 337 & 1207.55 & \textbf{1207.55} & 66.25 & \textbf{66.41} & 676 & 343.57 & \textbf{372.60} & 18.75 & \textbf{19.67} & 2088 \\
& & ONNX & - & - & - & - & - & - & - & - & - & - & - & - & - & - & - \\
& & LiteRT & - & - & - & - & - & - & - & - & - & - & - & - & - & - & - \\
\cmidrule(lr){1-18}
NPU & 32 & ET QNN & 1462.86 & \textbf{1542.17} & 61.02 & \textbf{62.38} & 681 & 2813.19 & \textbf{2976.74} & 46.50 & \textbf{46.57} & 1434 & 1161.29 & \textbf{1229.27} & 18.12 & \textbf{19.63} & 3584 \\
& - & QAIRT & - & - & - & - & - & 2277.90 & \textbf{2392.34} & 52.03 & \textbf{52.72} & 1229 & - & - & - & - & - \\
& 32 & llama.cpp & 343.10 & \textbf{409.10} & 22.50 & \textbf{23.40} & 442 & 329.90 & \textbf{374.40} & 23.60 & \textbf{25.60} & 728 & 114.40 & \textbf{130.00} & 12.10 & \textbf{12.50} & 2216 \\
& & LiteRT & - & - & - & - & - & - & - & - & - & - & - & - & - & - & - \\
\midrule
\multicolumn{18}{l}{\textit{Google Pixel 9 Pro XL (Tensor G4)}} \\
\midrule
CPU & 32 & ET XNNPACK & 169.13 & \textbf{299.54} & 37.07 & \textbf{40.38} & 417 & 235.51 & \textbf{245.92} & 31.29 & \textbf{32.99} & 821 & 39.32 & \textbf{51.94} & 7.93 & \textbf{11.25} & 2428 \\
& & llama.cpp & 240.60 & \textbf{241.10} & 46.50 & \textbf{46.60} & 442 & 158.10 & \textbf{202.40} & 29.60 & \textbf{30.90} & 728 & 57.20 & \textbf{58.60} & 10.00 & \textbf{10.30} & 2216 \\
& & ONNX & 144.42 & \textbf{259.93} & 26.52 & \textbf{35.23} & 376 & 133.25 & \textbf{156.72} & 22.55 & \textbf{27.07} & 769 & 32.75 & \textbf{34.89} & 6.06 & \textbf{6.48} & 2337 \\
& & LiteRT & 96.03 & \textbf{97.74} & 9.53 & \textbf{10.08} & 326 & 94.03 & \textbf{99.19} & 19.93 & \textbf{20.74} & 667 & - & - & - & - & - \\
\noalign{\vskip 3pt}
\hdashline
\noalign{\vskip 3pt}
& 128 & ET XNNPACK & 187.99 & \textbf{379.51} & 41.92 & \textbf{42.22} & 376 & 270.61 & \textbf{296.98} & 34.24 & \textbf{35.03} & 742 & 55.65 & \textbf{63.27} & 12.00 & \textbf{12.51} & 2201 \\
& & ONNX & 229.85 & \textbf{285.70} & 32.71 & \textbf{34.27} & 323 & 256.34 & \textbf{267.86} & 37.98 & \textbf{39.35} & 659 & 38.28 & \textbf{54.64} & 6.87 & \textbf{7.76} & 1994 \\
& & LiteRT & 101.14 & \textbf{109.68} & 10.22 & \textbf{10.31} & 299 & 117.00 & \textbf{117.51} & 21.48 & \textbf{21.65} & 612 & - & - & - & - & - \\
\cmidrule(lr){1-18}
GPU & 32 & ET Vulkan & 313.10 & \textbf{329.99} & 20.59 & \textbf{20.81} & 457 & 192.77 & \textbf{197.38} & 23.43 & \textbf{23.54} & 920 & 49.67 & \textbf{50.22} & 8.67 & \textbf{8.86} & 2829 \\
& & llama.cpp & 75.50 & \textbf{76.40} & 24.00 & \textbf{31.30} & 442 & 36.00 & \textbf{38.70} & 15.70 & \textbf{18.30} & 728 & 11.50 & \textbf{11.70} & 6.70 & \textbf{6.80} & 2216 \\
& & ONNX & - & - & - & - & - & - & - & - & - & - & - & - & - & - & - \\
& & LiteRT & - & - & - & - & - & - & - & - & - & - & - & - & - & - & - \\
\noalign{\vskip 3pt}
\hdashline
\noalign{\vskip 3pt}
& 128 & ET Vulkan & 591.01 & \textbf{601.83} & 20.71 & \textbf{21.12} & 337 & 530.02 & \textbf{540.08} & 23.74 & \textbf{24.03} & 676 & 119.64 & \textbf{120.07} & 9.58 & \textbf{9.64} & 2088 \\
& & ONNX & - & - & - & - & - & - & - & - & - & - & - & - & - & - & - \\
& & LiteRT & - & - & - & - & - & - & - & - & - & - & - & - & - & - & - \\
\bottomrule
\end{tabular}%
}
\end{table*}

\begin{table*}[t]
\centering
\small
\caption{ExecuTorch (ET) inference time in milliseconds (avg, p5, p95) for MV3, ResNet50, ViT, and Swin-T compared against other frameworks for different backends. The ``dtype'' column indicates the inference precision.}
\label{tab:vision_infer_time}
\resizebox{\textwidth}{!}{%
\begin{tabular}{llc rrr rrr rrr rrr}
\toprule
& & & \multicolumn{3}{c}{\textbf{MV3}} & \multicolumn{3}{c}{\textbf{ResNet50}} & \multicolumn{3}{c}{\textbf{ViT}} & \multicolumn{3}{c}{\textbf{Swin-T}} \\
\cmidrule(lr){4-6} \cmidrule(lr){7-9} \cmidrule(lr){10-12} \cmidrule(lr){13-15}
\textbf{Hardware} & \textbf{dtype} & \textbf{Framework} & \textbf{avg} & \textbf{p5} & \textbf{p95} & \textbf{avg} & \textbf{p5} & \textbf{p95} & \textbf{avg} & \textbf{p5} & \textbf{p95} & \textbf{avg} & \textbf{p5} & \textbf{p95} \\
\midrule
\multicolumn{15}{l}{\textit{Samsung Galaxy S25 Ultra (Snapdragon 8 Elite)}} \\
\midrule
CPU & int8 & ET XNNPACK & \textbf{0.51} & 0.46 & 0.72 & \textbf{4.95} & 4.54 & 5.38 & \textbf{64.97} & 53.31 & 76.19 & \textbf{23.06} & 21.06 & 26.36 \\
& & LiteRT & \textbf{0.70} & 0.66 & 0.83 & \textbf{8.99} & 8.86 & 9.27 & \textbf{115.91} & 113.77 & 131.01 & - & - & - \\
& & ONNX & \textbf{1.21} & 1.12 & 1.30 & \textbf{16.14} & 16.00 & 16.26 & \textbf{84.46} & 75.43 & 94.69 & \textbf{44.59} & 44.13 & 45.40 \\
\cmidrule(lr){1-15}
GPU & int8 & ET Vulkan & - & - & - & \textbf{5.44} & 4.87 & 5.65 & - & - & - & - & - & - \\
& & ONNX & \textbf{1.20} & 1.11 & 1.28 & - & - & - & \textbf{137.22} & 129.64 & 142.97 & - & - & - \\
\noalign{\vskip 3pt}
\hdashline
\noalign{\vskip 3pt}
& fp16 & ET Vulkan & \textbf{2.20} & 2.20 & 2.21 & \textbf{22.23} & 22.03 & 23.75 & \textbf{136.11} & 125.95 & 140.55 & \textbf{36.50} & 33.44 & 40.29 \\
& & LiteRT & \textbf{0.83} & 0.65 & 1.20 & - & - & - & \textbf{329.52} & 288.16 & 425.49 & - & - & - \\
& & ONNX & \textbf{1.82} & 1.50 & 2.94 & \textbf{6.47} & 6.01 & 7.09 & \textbf{168.54} & 129.25 & 241.39 & - & - & - \\
\cmidrule(lr){1-15}
NPU & int8 & ET QNN & \textbf{0.24} & 0.23 & 0.25 & \textbf{0.55} & 0.50 & 0.60 & \textbf{3.81} & 3.79 & 3.83 & \textbf{3.38} & 3.35 & 3.40 \\
& & LiteRT & - & - & - & \textbf{8.96} & 8.76 & 9.58 & \textbf{91.02} & 90.77 & 107.07 & - & - & - \\
& & ONNX & \textbf{7.78} & 7.71 & 7.88 & - & - & - & \textbf{175.94} & 173.50 & 178.69 & - & - & - \\
\midrule
\multicolumn{15}{l}{\textit{Google Pixel 9 Pro XL (Tensor G4)}} \\
\midrule
CPU & int8 & ET XNNPACK & \textbf{1.04} & 0.72 & 2.30 & \textbf{11.01} & 10.45 & 11.31 & \textbf{80.45} & 79.75 & 81.17 & \textbf{37.91} & 37.01 & 38.87 \\
& & LiteRT & - & - & - & \textbf{15.35} & 13.62 & 38.98 & \textbf{468.78} & 404.09 & 494.06 & - & - & - \\
& & ONNX & \textbf{1.91} & 1.69 & 2.60 & \textbf{15.44} & 15.21 & 15.59 & \textbf{103.82} & 89.95 & 175.27 & \textbf{41.48} & 41.25 & 41.71 \\
\cmidrule(lr){1-15}
GPU & int8 & ET Vulkan & - & - & - & \textbf{13.89} & 13.69 & 14.40 & - & - & - & - & - & - \\
\noalign{\vskip 3pt}
\hdashline
\noalign{\vskip 3pt}
& fp16 & ET Vulkan & \textbf{6.65} & 6.36 & 7.57 & \textbf{64.44} & 64.15 & 64.66 & \textbf{392.40} & 381.56 & 406.68 & \textbf{98.86} & 93.38 & 102.59 \\
\midrule
\multicolumn{15}{l}{\textit{Apple iPhone 15 Pro (A17 Pro)}} \\
\midrule
Auto & fp16 & ET CoreML & \textbf{0.40} & 0.35 & 0.47 & \textbf{1.60} & 1.57 & 1.63 & \textbf{10.55} & 10.49 & 10.65 & \textbf{8.70} & 8.40 & 9.13 \\
& & CoreML & \textbf{0.42} & 0.36 & 0.48 & \textbf{1.68} & 1.58 & 2.06 & \textbf{10.56} & 10.51 & 10.62 & - & - & - \\
\bottomrule
\end{tabular}%
}
\end{table*}

\textbf{Dense LLM Performance} - We benchmarked Qwen3 0.6B, Llama 3.2 1B, and
Phi4 Mini to showcase inference performance across a range of model sizes.
Quantization configurations were standardized as much as possible across
frameworks to ensure comparable model quality. For GPU and CPU inference on
ExecuTorch, ONNX, and LiteRT, model weights are group-wise quantized to 4-bit
precision and activations are dynamically quantized to 8-bit precision using
runtime-computed quantization parameters. Results for 32 and 128 quantization
group sizes are shown. For llama.cpp, the Q4\_0 quantization format is used;
most weights are quantized to 4-bit with a quantization group
size of 32. However, unlike the other frameworks, the LM-head weights are
quantized to 6 bits, and dynamic quantization of activations may or may not be
performed depending on the backend. For inference on Qualcomm NPU, QNN/Qualcomm
AI Runtime (QAIRT) requires that models be statically quantized, with 4-bit
weights and 16-bit activations. ExecuTorch's QNN delegate uses group-wise
quantized weights with a group size of 32, while QAIRT uses channel-wise
quantized weights. For NPU inference via llama.cpp, Q4\_0 quantization is used.

All models were benchmarked with 256 prompt tokens and 256 generated tokens; 3
runs were performed for each model, and the minimum/maximum throughput values
observed are recorded in Table~\ref{tab:llm_tps}. Models were configured to use
a maximum context length of 2048; for ExecuTorch, the maximum context length is
configured at export time, and for production settings higher values can be used
if needed. To mitigate the impact of thermal throttling, the device undergoes a
60-second cooldown period between runs. For CPU inference, the number of threads
is set to the number of performance cores on the device: 8 for the Samsung
Galaxy S25 Ultra and 4 for the Google Pixel 9 Pro XL. A warmup run is performed
before measurement to mitigate ``cold-start'' effects.

The model size reported for each framework is the size of the model artifacts
produced by each framework for a given model. ExecuTorch, LiteRT, and llama.cpp
all produce self-contained files (\texttt{.pte}, \texttt{.litertlm}, and
\texttt{.gguf} respectively) which contain the constant and weight tensor data
(i.e., quantized weights, quantization parameters, etc.) as well as serialized
model representations required for the framework to execute the model. ONNX uses
a \texttt{.onnx} file to store the model representation, and a separate
\texttt{.onnx.data} file to store constant and weight tensor data; the model
artifact size for ONNX is the sum of the sizes of these two files.

ExecuTorch's model artifacts tend to be larger compared to other frameworks. For
XNNPACK, this is because the delegate does not yet support tied embeddings (a
technique which shares the weight tensor between the embedding layer and the
LM-head linear layer), which results in a duplicated embedding table. For
Vulkan, although tied embeddings are supported, the delegate pre-computes
per-group integer weight sums (required for quantized accumulation) during
export and stores them in the model artifact. The overhead of these pre-computed
sums increases with smaller group sizes, which explains the difference in
model size between 32 and 128 group sizes. For both QNN and QAIRT, models use
16-bit embeddings and 8-bit LM-head, which prevents the use of tied embeddings.
QNN also contains higher quantization parameter overhead compared to QAIRT due
to the use of group-wise quantization.

ExecuTorch's XNNPACK delegate delivers strong performance compared to ONNX and
LiteRT across both devices. On the Samsung Galaxy S25, llama.cpp demonstrates
higher decode throughput for Qwen3 and Phi4 Mini (although prefill throughput
is comparable) because its attention implementation is more efficient for
single-token decode than ExecuTorch's.

ExecuTorch's Vulkan delegate is able to execute all models with full graph
delegation; no operators fall back to CPU. It underperforms llama.cpp in
prefill throughput on the Samsung Galaxy S25 Ultra, but tends to deliver better
token generation throughput. On the Pixel 9 Pro XL, the Vulkan delegate greatly
outperforms llama.cpp in prefill throughput, but this is because llama.cpp
currently only contains optimized compute shaders for Adreno GPUs. Although
group size has a large impact on prefill throughput for both devices, for the
Pixel 9 Pro XL, prefill throughput drops sharply at group size 32 compared to
128. This suggests a threshold at which the number of unique quantization
parameters fetched during the requantization step of quantized linear layers
increases memory traffic enough to cause GPU cache thrashing. For LiteRT, we
observed a segmentation fault when attempting to benchmark exported models with
GPU acceleration; for ONNX, we could not find a way to execute LLMs with
GPU acceleration.

For a detailed operator-level performance comparison of ExecuTorch and llama.cpp
across all 3 models on the Samsung Galaxy S25 Ultra, see
Appendix~\ref{sec:llm-breakdown}. Note that the analysis only covers CPU and GPU
inference.

ExecuTorch's QNN delegate demonstrates stronger prefill throughput compared to
executing via QAIRT for Llama 3.2 1B. QAIRT achieves higher
decode throughput, which may be due to its use of per-channel quantization
rather than the per-group quantization used by ExecuTorch's QNN delegate. We
could not generate QAIRT binaries for Qwen3 0.6B and Phi4 Mini due to errors
during the model export process. llama.cpp's Hexagon backend (currently marked
experimental) targets NPU inference using custom DSP kernels, and some ops may
be falling back to the CPU, which would explain the much lower throughput
compared to QNN. In contrast, the QNN delegate executes the model with full
graph delegation with no operators falling back to CPU.

\textbf{Vision Model Performance} - Performance results for MV3, ResNet50, ViT,
and Swin-T are reported in Table~\ref{tab:vision_infer_time}. These models are
selected as a representative sample of convolution-based and transformer-based
image processing workloads. Each model was benchmarked with 10 warmup iterations
and 200 inference iterations, and the average, p5, and p95 inference latencies
in milliseconds are reported. For CPU inference, the number of threads is set to
the number of performance cores on the device; 8 for the Samsung Galaxy S25
Ultra and 4 for the Google Pixel 9 Pro XL.

We encountered errors when exporting the Swin-T model with LiteRT, so no
measurements are reported for Swin-T on LiteRT. We were also unable
to collect GPU inference measurements for many LiteRT models due to a segmentation
fault that occurred after loading the GPU acceleration library. For ONNX,
GPU/NPU inference was tested via the QNN execution provider, which is not
available for the Pixel 9 Pro XL. A runtime exception was encountered when
executing Swin-T with the QNN execution provider, and therefore no data was
collected for that model on GPU/NPU with ONNX.

For CPU inference, ExecuTorch's XNNPACK delegate delivers extremely strong
performance relative to LiteRT and ONNX.

For GPU inference, ExecuTorch's Vulkan delegate executes MobileNet V3
and ResNet50 (both quantized and fp16 variants) with full graph delegation. For
ViT-B/16, 4 unsupported operator types (72 instances total) in the attention
masking pipeline---\texttt{mul.Scalar}, \texttt{logical\_not},
\texttt{eq.Scalar}, and \texttt{any.dim}---cause the model to be split into 25
Vulkan partitions. For Swin-T, 7 unsupported operator types ($\sim$160
instances)---primarily \texttt{slice\_scatter}, \texttt{fmod.Scalar}, \texttt{index.Tensor}
with 2D sources---produce 12 partitions. Although CPU fallback operators account
for $\sim$29\% of ViT execution latency and $\sim$22\% of Swin-T execution latency, the
overhead introduced by graph breaks (i.e., copying tensors between CPU and GPU)
accounts for only 5\%--6\% of execution latency.

Generally, the delegate delivers comparable performance to other
frameworks on the Samsung Galaxy S25. A notable exception is ResNet50, where
ONNX achieves much faster inference. Likewise, LiteRT achieves much faster
inference on MobileNet V3 compared to both ExecuTorch and ONNX. These gaps do not
appear to be consistent across models. Since neither QNN nor
LiteRT's GPU acceleration library is open source, it is difficult to
diagnose the source of the performance gap. For int8 inference on the GPU,
support for static int8 quantization in the Vulkan delegate is an ongoing
effort, and so far only ResNet50 is supported among the models tested.

For NPU inference, ExecuTorch's QNN delegate delivers extremely strong
performance relative to LiteRT and ONNX. We found that for LiteRT and ONNX, the
NPU execution provider / accelerator was only claiming a limited number of
nodes in the model graph, resulting in a majority of model inference being
executed on the CPU.

On the iPhone 15 Pro, ExecuTorch's CoreML delegate matches or slightly
outperforms native CoreML across all four models, and is the only configuration
that produces results for Swin-T. These results confirm that ExecuTorch's
delegation overhead is negligible compared to directly running CoreML.

\section{Limitations and Future Work}

\textbf{Model exportability:} ExecuTorch relies on \texttt{torch.export} for
graph capture. Unfortunately, there are several classes of models that pose
challenges for export.

\begin{compactitem}
\item \textbf{Data-dependent control flow:} models that contain operations where
control flow depends on runtime tensor values, e.g., models with dynamic padding,
data-dependent slicing, or models that branch on data-dependent values (e.g., beam
search).

\item \textbf{Dynamic Shapes:} models with input-dependent tensor dimensions
such as dynamic LSTMs or Mask R-CNN architectures produce graphs whose structure
changes at runtime. These often require splitting the model into separately
exported subgraphs or rewriting the model to be export-friendly.

\item \textbf{Custom operators:} Models that rely on custom C++ or CUDA kernels
(e.g., FlashAttention) require those kernels to be registered with a fake-tensor implementation
that describes output shapes symbolically. A corresponding kernel must also be
registered in the ExecuTorch runtime, and each target delegate must implement support for
it, in order for ExecuTorch to handle the custom operator.

\end{compactitem}

Models containing any of the above elements may require changes such as:

\begin{compactitem}
\item Rewriting control flow with higher-order operators such as
\texttt{torch.cond} (if/else), \texttt{torch.scan} or
\texttt{torch.while\_loop} (loops), and \texttt{torch.where} (element-wise
selection) \cite{wu2025control}.

\item Wrapping untraceable ops in a custom op.

\item Adding assertions (\texttt{torch.\_check}), which serve as compiler hints
for constraining dynamic shapes.
\end{compactitem}

\textbf{Hardware retargetability:} AOT compilation optimizes models for
specific hardware, but hardware diversity in the Android ecosystem (e.g.,
Qualcomm, MediaTek, Samsung NPUs) may require developers to query device
capabilities and select a hardware-appropriate model file when downloading
models from a delivery service, or bundle multiple hardware-specific model files
in one APK (perhaps with a common shared weight file). This increases
engineering complexity compared to runtime-retargetable solutions (like ONNX or
LiteRT), which are more flexible but forgo AOT optimizations.

\textbf{Desktop/Laptop Support:} ExecuTorch accelerates inference on consumer
desktops and laptops using backends like XNNPACK, OpenVINO, and QNN. With rising
demand for local inference (exemplified by llama.cpp, MLX), ExecuTorch is now
experimenting with CUDA and Metal backend support, leveraging PyTorch’s
AOTInductor technology.

\textbf{Sparsity:} ExecuTorch’s IR can represent sparse weights using dense
values and mask tensors, but hardware-accelerated sparse kernels (e.g., 2:4
structured sparsity) are not yet widely available on edge targets. Sparsity
support remains a future direction as backend capabilities mature.




\section*{Acknowledgements}

The completion of this project would not have been possible without the support,
input, and contributions of numerous individuals and institutions. We wish to
express our sincere gratitude to all those who offered their expertise,
resources, and guidance throughout this project. For a full list, please see
Appendix~\ref{sec:ack-appendix}.


\nocite{langley00}

\FloatBarrier
\clearpage
\bibliography{example_paper}

@inproceedings{langley00,
 author    = {P. Langley},
 title     = {Crafting Papers on Machine Learning},
 year      = {2000},
 pages     = {1207--1216},
 editor    = {Pat Langley},
 booktitle     = {Proceedings of the 17th International Conference
              on Machine Learning (ICML 2000)},
 address   = {Stanford, CA},
 publisher = {Morgan Kaufmann}
}

@inproceedings{pytorch2,
author = {Ansel, Jason and Yang, Edward and He, Horace and Gimelshein, Natalia and Jain, Animesh and Voznesensky, Michael and Bao, Bin and Bell, Peter and Berard, David and Burovski, Evgeni and Chauhan, Geeta and Chourdia, Anjali and Constable, Will and Desmaison, Alban and DeVito, Zachary and Ellison, Elias and Feng, Will and Gong, Jiong and Gschwind, Michael and Hirsh, Brian and Huang, Sherlock and Kalambarkar, Kshiteej and Kirsch, Laurent and Lazos, Michael and Lezcano, Mario and Liang, Yanbo and Liang, Jason and Lu, Yinghai and Luk, C. K. and Maher, Bert and Pan, Yunjie and Puhrsch, Christian and Reso, Matthias and Saroufim, Mark and Siraichi, Marcos Yukio and Suk, Helen and Zhang, Shunting and Suo, Michael and Tillet, Phil and Zhao, Xu and Wang, Eikan and Zhou, Keren and Zou, Richard and Wang, Xiaodong and Mathews, Ajit and Wen, William and Chanan, Gregory and Wu, Peng and Chintala, Soumith},
title = {PyTorch 2: Faster Machine Learning Through Dynamic Python Bytecode Transformation and Graph Compilation},
year = {2024},
isbn = {9798400703850},
publisher = {Association for Computing Machinery},
address = {New York, NY, USA},
url = {https://doi.org/10.1145/3620665.3640366},
doi = {10.1145/3620665.3640366},
booktitle = {Proceedings of the 29th ACM International Conference on Architectural Support for Programming Languages and Operating Systems, Volume 2},
pages = {929–947},
numpages = {19},
location = {La Jolla, CA, USA},
series = {ASPLOS '24}
}

@misc{torch_fx,
  author       = {James K. Reed and
                  Zachary DeVito and
                  Horace He and
                  Ansley Ussery and
                  Jason Ansel},
  title        = {torch.fx: Practical Program Capture and Transformation for Deep Learning
                  in Python},
  journal      = {CoRR},
  volume       = {abs/2112.08429},
  year         = {2021},
  url          = {https://arxiv.org/abs/2112.08429},
  eprinttype    = {arXiv},
  eprint       = {2112.08429},
  bibsource    = {dblp computer science bibliography, https://dblp.org}
}

@misc{core_aten_ir,
  title = {IRs},
  author = {{PyTorch Developers}},
  year = {2022},
  url = {https://docs.pytorch.org/docs/2.9/torch.compiler_ir.html},
}

@misc{torch_export,
  title = {torch.export},
  author = {{PyTorch Developers}},
  year = {2022},
  url = {https://docs.pytorch.org/docs/2.9/export.html},
}

@misc{pt2e_ptq,
  title = {PyTorch 2 Export Post Training Quantization},
  author = {{Jerry Zhang}},
  year = {2025},
  url = {https://docs.pytorch.org/ao/stable/tutorials_source/pt2e_quant_ptq.html},
}

@misc{pt2e_qat,
  title = {PyTorch 2 Export Quantization-Aware Training (QAT)},
  author = {{Andrew Or}},
  year = {2025},
  url = {https://docs.pytorch.org/ao/stable/tutorials_source/pt2e_quant_qat.html},
}

@misc{pytorch_tensor_subclass,
  title = {Subclassing torch.Tensor},
  author = {{PyTorch}},
  year = {2025},
  url = {https://docs.pytorch.org/docs/stable/notes/extending.html#subclassing-torch-tensor},
}

@misc{mobile_interpreter,
  title = {PyTorch 1.9 Release, including torch.linalg and Mobile Interpreter},
  author = {{PyTorch Foundation}},
  year = {2021},
  url = {https://pytorch.org/blog/pytorch-1-9-released/},
}

@misc{onnx,
    author = {Bai, Junjie and Lu, Fang and Zhang, Ke and others},
    title = {ONNX: Open Neural Network Exchange},
    year = {2019},
    publisher = {GitHub},
    journal = {GitHub repository},
    howpublished = {\url{https://github.com/onnx/onnx}},
    commit = {94d238d96e3fb3a7ba34f03c284b9ad3516163be}
}

@misc{mlx2023,
  author = {Awni Hannun and Jagrit Digani and Angelos Katharopoulos and Ronan Collobert},
  title = {{MLX}: Efficient and flexible machine learning on Apple silicon},
  url = {https://github.com/ml-explore},
  version = {0.0},
  year = {2023},
}

@misc{huggingface_optimum_executorch,
  author       = {{PyTorch Developers}},
  title        = {{Optimum ExecuTorch} — Optimize and deploy Hugging Face models with ExecuTorch},
  howpublished = {\url{https://github.com/huggingface/optimum-executorch}},
  year         = {2025},
  note         = {Accessed: 2025-10-28}
}

@misc{yang2025qwen3technicalreport,
      title={Qwen3 Technical Report}, 
      author={An Yang and Anfeng Li and Baosong Yang and Beichen Zhang and Binyuan Hui and Bo Zheng and Bowen Yu and Chang Gao and Chengen Huang and Chenxu Lv and Chujie Zheng and Dayiheng Liu and Fan Zhou and Fei Huang and Feng Hu and Hao Ge and Haoran Wei and Huan Lin and Jialong Tang and Jian Yang and Jianhong Tu and Jianwei Zhang and Jianxin Yang and Jiaxi Yang and Jing Zhou and Jingren Zhou and Junyang Lin and Kai Dang and Keqin Bao and Kexin Yang and Le Yu and Lianghao Deng and Mei Li and Mingfeng Xue and Mingze Li and Pei Zhang and Peng Wang and Qin Zhu and Rui Men and Ruize Gao and Shixuan Liu and Shuang Luo and Tianhao Li and Tianyi Tang and Wenbiao Yin and Xingzhang Ren and Xinyu Wang and Xinyu Zhang and Xuancheng Ren and Yang Fan and Yang Su and Yichang Zhang and Yinger Zhang and Yu Wan and Yuqiong Liu and Zekun Wang and Zeyu Cui and Zhenru Zhang and Zhipeng Zhou and Zihan Qiu},
      year={2025},
      eprint={2505.09388},
      archivePrefix={arXiv},
      primaryClass={cs.CL},
      url={https://arxiv.org/abs/2505.09388}, 
}

@misc{liu2025spinquantllmquantizationlearned,
      title={SpinQuant: LLM quantization with learned rotations}, 
      author={Zechun Liu and Changsheng Zhao and Igor Fedorov and Bilge Soran and Dhruv Choudhary and Raghuraman Krishnamoorthi and Vikas Chandra and Yuandong Tian and Tijmen Blankevoort},
      year={2025},
      eprint={2405.16406},
      archivePrefix={arXiv},
      primaryClass={cs.LG},
      url={https://arxiv.org/abs/2405.16406}, 
}

@misc{lin2024awqactivationawareweightquantization,
      title={AWQ: Activation-aware Weight Quantization for LLM Compression and Acceleration}, 
      author={Ji Lin and Jiaming Tang and Haotian Tang and Shang Yang and Wei-Ming Chen and Wei-Chen Wang and Guangxuan Xiao and Xingyu Dang and Chuang Gan and Song Han},
      year={2024},
      eprint={2306.00978},
      archivePrefix={arXiv},
      primaryClass={cs.CL},
      url={https://arxiv.org/abs/2306.00978}, 
}

@misc{jia2014caffeconvolutionalarchitecturefast,
      title={Caffe: Convolutional Architecture for Fast Feature Embedding}, 
      author={Yangqing Jia and Evan Shelhamer and Jeff Donahue and Sergey Karayev and Jonathan Long and Ross Girshick and Sergio Guadarrama and Trevor Darrell},
      year={2014},
      eprint={1408.5093},
      archivePrefix={arXiv},
      primaryClass={cs.CV},
      url={https://arxiv.org/abs/1408.5093}, 
}

@misc{liu2025voxtral,
      title={Voxtral}, 
      author={Alexander H. Liu and Andy Ehrenberg and Andy Lo and Clément Denoix and Corentin Barreau and Guillaume Lample and Jean-Malo Delignon and Khyathi Raghavi Chandu and Patrick von Platen and Pavankumar Reddy Muddireddy and Sanchit Gandhi and Soham Ghosh and Srijan Mishra and Thomas Foubert and Abhinav Rastogi and Adam Yang and Albert Q. Jiang and Alexandre Sablayrolles and Amélie Héliou and Amélie Martin and Anmol Agarwal and Antoine Roux and Arthur Darcet and Arthur Mensch and Baptiste Bout and Baptiste Rozière and Baudouin De Monicault and Chris Bamford and Christian Wallenwein and Christophe Renaudin and Clémence Lanfranchi and Darius Dabert and Devendra Singh Chaplot and Devon Mizelle and Diego de las Casas and Elliot Chane-Sane and Emilien Fugier and Emma Bou Hanna and Gabrielle Berrada and Gauthier Delerce and Gauthier Guinet and Georgii Novikov and Guillaume Martin and Himanshu Jaju and Jan Ludziejewski and Jason Rute and Jean-Hadrien Chabran and Jessica Chudnovsky and Joachim Studnia and Joep Barmentlo and Jonas Amar and Josselin Somerville Roberts and Julien Denize and Karan Saxena and Karmesh Yadav and Kartik Khandelwal and Kush Jain and Lélio Renard Lavaud and Léonard Blier and Lingxiao Zhao and Louis Martin and Lucile Saulnier and Luyu Gao and Marie Pellat and Mathilde Guillaumin and Mathis Felardos and Matthieu Dinot and Maxime Darrin and Maximilian Augustin and Mickaël Seznec and Neha Gupta and Nikhil Raghuraman and Olivier Duchenne and Patricia Wang and Patryk Saffer and Paul Jacob and Paul Wambergue and Paula Kurylowicz and Philomène Chagniot and Pierre Stock and Pravesh Agrawal and Rémi Delacourt and Romain Sauvestre and Roman Soletskyi and Sagar Vaze and Sandeep Subramanian and Saurabh Garg and Shashwat Dalal and Siddharth Gandhi and Sumukh Aithal and Szymon Antoniak and Teven Le Scao and Thibault Schueller and Thibaut Lavril and Thomas Robert and Thomas Wang and Timothée Lacroix and Tom Bewley and Valeriia Nemychnikova and Victor Paltz and Virgile Richard and Wen-Ding Li and William Marshall and Xuanyu Zhang and Yihan Wan and Yunhao Tang},
      year={2025},
      eprint={2507.13264},
      archivePrefix={arXiv},
      primaryClass={cs.SD},
      url={https://arxiv.org/abs/2507.13264}, 
}

@misc{gemmateam2025gemma3technicalreport,
      title={Gemma 3 Technical Report}, 
      author={Gemma Team and Aishwarya Kamath and Johan Ferret and Shreya Pathak and Nino Vieillard and Ramona Merhej and Sarah Perrin and Tatiana Matejovicova and Alexandre Ramé and Morgane Rivière and Louis Rouillard and Thomas Mesnard and Geoffrey Cideron and Jean-bastien Grill and Sabela Ramos and Edouard Yvinec and Michelle Casbon and Etienne Pot and Ivo Penchev and Gaël Liu and Francesco Visin and Kathleen Kenealy and Lucas Beyer and Xiaohai Zhai and Anton Tsitsulin and Robert Busa-Fekete and Alex Feng and Noveen Sachdeva and Benjamin Coleman and Yi Gao and Basil Mustafa and Iain Barr and Emilio Parisotto and David Tian and Matan Eyal and Colin Cherry and Jan-Thorsten Peter and Danila Sinopalnikov and Surya Bhupatiraju and Rishabh Agarwal and Mehran Kazemi and Dan Malkin and Ravin Kumar and David Vilar and Idan Brusilovsky and Jiaming Luo and Andreas Steiner and Abe Friesen and Abhanshu Sharma and Abheesht Sharma and Adi Mayrav Gilady and Adrian Goedeckemeyer and Alaa Saade and Alex Feng and Alexander Kolesnikov and Alexei Bendebury and Alvin Abdagic and Amit Vadi and András György and André Susano Pinto and Anil Das and Ankur Bapna and Antoine Miech and Antoine Yang and Antonia Paterson and Ashish Shenoy and Ayan Chakrabarti and Bilal Piot and Bo Wu and Bobak Shahriari and Bryce Petrini and Charlie Chen and Charline Le Lan and Christopher A. Choquette-Choo and CJ Carey and Cormac Brick and Daniel Deutsch and Danielle Eisenbud and Dee Cattle and Derek Cheng and Dimitris Paparas and Divyashree Shivakumar Sreepathihalli and Doug Reid and Dustin Tran and Dustin Zelle and Eric Noland and Erwin Huizenga and Eugene Kharitonov and Frederick Liu and Gagik Amirkhanyan and Glenn Cameron and Hadi Hashemi and Hanna Klimczak-Plucińska and Harman Singh and Harsh Mehta and Harshal Tushar Lehri and Hussein Hazimeh and Ian Ballantyne and Idan Szpektor and Ivan Nardini and Jean Pouget-Abadie and Jetha Chan and Joe Stanton and John Wieting and Jonathan Lai and Jordi Orbay and Joseph Fernandez and Josh Newlan and Ju-yeong Ji and Jyotinder Singh and Kat Black and Kathy Yu and Kevin Hui and Kiran Vodrahalli and Klaus Greff and Linhai Qiu and Marcella Valentine and Marina Coelho and Marvin Ritter and Matt Hoffman and Matthew Watson and Mayank Chaturvedi and Michael Moynihan and Min Ma and Nabila Babar and Natasha Noy and Nathan Byrd and Nick Roy and Nikola Momchev and Nilay Chauhan and Noveen Sachdeva and Oskar Bunyan and Pankil Botarda and Paul Caron and Paul Kishan Rubenstein and Phil Culliton and Philipp Schmid and Pier Giuseppe Sessa and Pingmei Xu and Piotr Stanczyk and Pouya Tafti and Rakesh Shivanna and Renjie Wu and Renke Pan and Reza Rokni and Rob Willoughby and Rohith Vallu and Ryan Mullins and Sammy Jerome and Sara Smoot and Sertan Girgin and Shariq Iqbal and Shashir Reddy and Shruti Sheth and Siim Põder and Sijal Bhatnagar and Sindhu Raghuram Panyam and Sivan Eiger and Susan Zhang and Tianqi Liu and Trevor Yacovone and Tyler Liechty and Uday Kalra and Utku Evci and Vedant Misra and Vincent Roseberry and Vlad Feinberg and Vlad Kolesnikov and Woohyun Han and Woosuk Kwon and Xi Chen and Yinlam Chow and Yuvein Zhu and Zichuan Wei and Zoltan Egyed and Victor Cotruta and Minh Giang and Phoebe Kirk and Anand Rao and Kat Black and Nabila Babar and Jessica Lo and Erica Moreira and Luiz Gustavo Martins and Omar Sanseviero and Lucas Gonzalez and Zach Gleicher and Tris Warkentin and Vahab Mirrokni and Evan Senter and Eli Collins and Joelle Barral and Zoubin Ghahramani and Raia Hadsell and Yossi Matias and D. Sculley and Slav Petrov and Noah Fiedel and Noam Shazeer and Oriol Vinyals and Jeff Dean and Demis Hassabis and Koray Kavukcuoglu and Clement Farabet and Elena Buchatskaya and Jean-Baptiste Alayrac and Rohan Anil and Dmitry and Lepikhin and Sebastian Borgeaud and Olivier Bachem and Armand Joulin and Alek Andreev and Cassidy Hardin and Robert Dadashi and Léonard Hussenot},
      year={2025},
      eprint={2503.19786},
      archivePrefix={arXiv},
      primaryClass={cs.CL},
      url={https://arxiv.org/abs/2503.19786}, 
}

@INPROCEEDINGS{9190246,
  author={Gadzicki, Konrad and Khamsehashari, Razieh and Zetzsche, Christoph},
  booktitle={2020 IEEE 23rd International Conference on Information Fusion (FUSION)}, 
  title={Early vs Late Fusion in Multimodal Convolutional Neural Networks}, 
  year={2020},
  volume={},
  number={},
  pages={1-6},
  doi={10.23919/FUSION45008.2020.9190246}
}

@inproceedings{wolf-etal-2020-transformers,
    title = "Transformers: State-of-the-Art Natural Language Processing",
    author = "Thomas Wolf and Lysandre Debut and Victor Sanh and Julien Chaumond and Clement Delangue and Anthony Moi and Pierric Cistac and Tim Rault and Rémi Louf and Morgan Funtowicz and Joe Davison and Sam Shleifer and Patrick von Platen and Clara Ma and Yacine Jernite and Julien Plu and Canwen Xu and Teven Le Scao and Sylvain Gugger and Mariama Drame and Quentin Lhoest and Alexander M. Rush",
    booktitle = "Proceedings of the 2020 Conference on Empirical Methods in Natural Language Processing: System Demonstrations",
    month = oct,
    year = "2020",
    address = "Online",
    publisher = "Association for Computational Linguistics",
    url = "https://www.aclweb.org/anthology/2020.emnlp-demos.6",
    pages = "38--45"
}

@misc{xnnpack,
  title = {XNNPACK: High-efficiency floating-point neural network inference operators for mobile and server platforms},
  author = {Google},
  howpublished = {\url{https://github.com/google/XNNPACK}},
  year = {2019},
  note = {Accessed: 2025-10-28}
}

@misc{kleidiai,
  title        = {KleidiAI: Open-source micro-kernel library for AI workloads on Arm CPUs},
  author       = {{Arm Ltd.}},
  howpublished = {\url{https://gitlab.arm.com/kleidi/kleidiai} (mirror: \url{https://github.com/ARM-software/kleidiai})},
  year         = {2024},
  note         = {Accessed: 2025-10-28},
}

@misc{coreml,
  title        = {Core ML: Machine Learning Framework for Apple Platforms},
  author       = {{Apple Inc.}},
  howpublished = {\url{https://developer.apple.com/documentation/coreml}},
  year         = {2023},
  note         = {Accessed: 2025-10-28}
}

@misc{tosa_spec,
  title        = {Tensor Operator Set Architecture (TOSA) Specification v1.0.1},
  author       = {ML Platform Consortium},
  howpublished = {\url{https://www.mlplatform.org/tosa/tosa_spec.html}},
  year         = {2023},
  note         = {Accessed: 2025-10-28}
}

@misc{arm_ethosu,
  title        = {Arm Ethos-U Ecosystem: MicroNPUs and Software for Efficient Edge AI},
  author       = {{Arm Ltd.}},
  howpublished = {\url{https://developer.arm.com/Processors/Ethos-U}},
  year         = {2024},
  note         = {Accessed: 2025-10-28}
}

@misc{vulkan_spec,
  title        = {Vulkan API Specification, Version 1.3},
  author       = {{Khronos Group}},
  institution  = {The Khronos Group Inc.},
  year         = {2023},
  url          = {https://registry.khronos.org/vulkan/specs/1.3/html/vkspec.html},
  note         = {Accessed: 2025-10-28}
}

@misc{qualcomm_ai_engine_direct,
  title        = {Qualcomm AI Engine Direct SDK},
  author       = {Qualcomm Technologies, Inc.},
  howpublished = {\url{https://www.qualcomm.com/developer/software/qualcomm-ai-engine-direct-sdk}},
  year         = {2024},
  note         = {Accessed: 2025-10-28}
}

@misc{wang2025optimizing,
  author        = {Wang, Xubin and Jia, Weijia},
  title         = {Optimizing Edge {AI}: A Comprehensive Survey on Data, Model, and System Strategies},
  year          = {2025},
  eprint        = {2501.03265},
  archivePrefix = {arXiv},
  primaryClass  = {cs.LG},
  url           = {https://arxiv.org/abs/2501.03265},
  doi           = {10.48550/arXiv.2501.03265}
}

@misc{wang2025deploying,
  author    = {Wang, Tianyu and Guo, Jinyang and Zhang, Bowen and Yang, Ge and Li, Dong},
  title     = {Deploying {AI} on Edge: Advancement and Challenges in Edge Intelligence},
  journal   = {Mathematics},
  volume    = {13},
  number    = {11},
  pages     = {1878},
  year      = {2025},
  publisher = {MDPI},
  doi       = {10.3390/math13111878},
  issn      = {2227-7390}
}

@misc{ng2025development,
  author    = {Ng, Madelena Y. and Helzer, Jarrod and Pfeffer, Michael A. and Seto, Tina and Hernandez-Boussard, Tina},
  title     = {Development of secure infrastructure for advancing generative artificial intelligence research in healthcare at an academic medical center},
  journal   = {Journal of the American Medical Informatics Association},
  volume    = {32},
  number    = {3},
  pages     = {586--588},
  year      = {2025},
  month     = {March},
  doi       = {10.1093/jamia/ocaf005},
  issn      = {1527-974X},
  publisher = {Oxford University Press}
}

@misc{kuo2020privacy,
  author    = {Kuo, Tsung-Ting and Kim, Jihoon and Gabriel, Rodney A.},
  title     = {Privacy-preserving model learning on a blockchain network-of-networks},
  journal   = {Journal of the American Medical Informatics Association},
  volume    = {27},
  number    = {3},
  pages     = {343--354},
  year      = {2020},
  month     = {March},
  doi       = {10.1093/jamia/ocz214},
  pmid      = {31943009},
  pmcid     = {PMC7025358},
  publisher = {Oxford University Press}
}

@misc{xu2021federated,
  author    = {Xu, Jie and Glicksberg, Benjamin S. and Su, Chang and Walker, Peter and Bian, Jiang and Wang, Fei},
  title     = {Federated Learning for Healthcare Informatics},
  journal   = {Journal of Healthcare Informatics Research},
  volume    = {5},
  number    = {1},
  pages     = {1--19},
  year      = {2021},
  month     = {March},
  doi       = {10.1007/s41666-020-00082-4},
  pmid      = {33204939},
  pmcid     = {PMC7659898},
  publisher = {Springer}
}

@inproceedings{sperling2024reducing,
  author    = {Sperling, N. and Ernst, R.},
  title     = {Reducing Communication Cost and Latency in Autonomous Vehicles with Subscriber-centric Selective Data Distribution},
  booktitle = {2024 IEEE 99th Vehicular Technology Conference (VTC Spring)},
  year      = {2024},
  address   = {Singapore},
  month     = {June},
  note      = {Document ID: 10683426}
}

@misc{nigade2024inference,
  author    = {Nigade, Vinod and Bauszat, Pablo and Bal, Henri E. and Wang, Lin},
  title     = {Inference serving with end-to-end latency {SLO}s over dynamic edge networks},
  journal   = {Real-Time Systems},
  volume    = {60},
  pages     = {239--290},
  year      = {2024},
  publisher = {Springer},
  doi       = {10.1007/s11241-024-09418-4}
}

@inproceedings{kang2024rtswap,
  author    = {Kang, Woosung and Lee, Jinkyu and Lee, Youngmoon and Oh, Sangeun and Lee, Kilho and Chwa, Hoon Sung},
  title     = {{RT-Swap}: Addressing {GPU} Memory Bottlenecks for Real-Time Multi-{DNN} Inference},
  booktitle = {2024 IEEE 30th Real-Time and Embedded Technology and Applications Symposium (RTAS)},
  year      = {2024},
  pages     = {373--385},
  doi       = {10.1109/RTAS61025.2024.00037},
  address   = {Hong Kong, China},
  month     = may
}

@misc{pons2023utilization,
  author  = {Pons, Mario and Valenzuela, Estuardo and Rodr{\\'i}guez, Brandon and Nolazco-Flores, Juan Arturo and Del-Valle-Soto, Carolina},
  title   = {Utilization of {5G} Technologies in {IoT} Applications: Current Limitations by Interference and Network Optimization Difficulties---{A} Review},
  journal = {Sensors},
  volume  = {23},
  number  = {8},
  pages   = {3876},
  year    = {2023},
  month   = apr,
  doi     = {10.3390/s23083876},
  issn    = {1424-8220}
}

@inproceedings{vasu2023fastvit,
  author    = {Vasu, Pavan Kumar Anasosalu and Gabriel, James and Zhu, Jeff and Tuzel, Oncel and Ranjan, Anurag},
  title     = {{FastViT}: A Fast Hybrid Vision Transformer Using Structural Reparameterization},
  booktitle = {Proceedings of the IEEE/CVF International Conference on Computer Vision (ICCV)},
  month     = {October},
  year      = {2023},
  pages     = {5785--5795}
}

@inproceedings{vasu2024mobileclip,
  author    = {Vasu, Pavan Kumar Anasosalu and Pouransari, Hadi and Faghri, Fartash and Vemulapalli, Raviteja and Tuzel, Oncel},
  title     = {MobileCLIP: Fast Image-Text Models through Multi-Modal Reinforced Training},
  booktitle = {Proceedings of the IEEE/CVF Conference on Computer Vision and Pattern Recognition (CVPR)},
  month     = {June},
  year      = {2024},
  pages     = {15963--15974}
}

@misc{ahsan2025hardware,
  author    = {Ahsan, S. M. Mojahidul and Hoque, Tamzidul and Hasan, Md Sakib and Chowdhury, Mrittika and Dhungel, Anurag},
  title     = {Hardware Accelerators for Artificial Intelligence},
  booktitle = {AI-Enabled Electronic Circuit and System Design},
  editor    = {Iranmanesh, A. and Sayadi, H.},
  year      = {2025},
  publisher = {Springer},
  address   = {Cham},
  pages     = {497--535},
  doi       = {10.1007/978-3-031-71436-8_14},
  isbn      = {978-3-031-71436-8}
}

@inproceedings{david2021tensorflowlite,
  author    = {David, Robert and Duke, Jared and Jain, Advait and Reddi, Vijay Janapa and Jeffries, Nat and Li, Jian and Kreeger, Nick and Nappier, Ian and Natraj, Meghna and Regev, Shlomi and Rhodes, Rocky and Wang, Tiezhen and Warden, Pete},
  title     = {{TensorFlow Lite Micro: Embedded Machine Learning on TinyML Systems}},
  booktitle = {Proceedings of Machine Learning and Systems},
  volume    = {3},
  pages     = {800--811},
  year      = {2021},
  editor    = {Smola, A. and Dimakis, A. and Stoica, I.},
  address   = {San Jose, CA, USA},
  month     = apr,
  url       = {https://proceedings.mlsys.org/paper_files/paper/2021/file/6c44dc73014d66ba49b28d483a8f8b0d-Paper.pdf},
  note      = {Now known as LiteRT}
}

@misc{apple2017coreml,
  title        = {{Core ML}},
  author       = {{Apple Inc.}},
  year         = {2017},
  organization = {Apple Inc.},
  address      = {Cupertino, CA},
  url          = {https://developer.apple.com/documentation/coreml},
  note         = {Machine learning framework for iOS, macOS, watchOS, and tvOS. Introduced at WWDC 2017}
}

@misc{onnxruntime2018,
  author       = {{Microsoft}},
  title        = {{ONNX Runtime}: Cross-platform, High Performance ML Inferencing and Training Accelerator},
  year         = {2018},
  month        = dec,
  howpublished = {\url{https://github.com/microsoft/onnxruntime}},
  url          = {https://onnxruntime.ai/},
  note         = {Open source inference engine}
}

@inproceedings{chen2018tvm,
  author    = {Tianqi Chen and Thierry Moreau and Ziheng Jiang and Lianmin Zheng and Eddie Yan and Haichen Shen and Meghan Cowan and Leyuan Wang and Yuwei Hu and Luis Ceze and Carlos Guestrin and Arvind Krishnamurthy},
  title     = {{TVM}: An Automated End-to-End Optimizing Compiler for Deep Learning},
  booktitle = {13th USENIX Symposium on Operating Systems Design and Implementation (OSDI 18)},
  year      = {2018},
  month     = oct,
  pages     = {578--594},
  address   = {Carlsbad, CA, USA},
  isbn      = {978-1-939133-08-3},
  url       = {https://www.usenix.org/conference/osdi18/presentation/chen},
  publisher = {USENIX Association}
}

@inproceedings{jiang2020mnn,
  author    = {Jiang, Xiaotang and Wang, Huan and Chen, Yiliu and Wu, Ziqi and Wang, Lichuan and Zou, Bin and Yang, Yafeng and Cui, Zongyang and Cai, Yu and Yu, Tianhang and Lv, Chengfei and Wu, Zhihua},
  title     = {{MNN}: A Universal and Efficient Inference Engine},
  booktitle = {Proceedings of Machine Learning and Systems},
  year      = {2020},
  volume    = {2},
  pages     = {1--13},
  url       = {https://proceedings.mlsys.org/paper_files/paper/2020/hash/bc19061f88f16e9ed4a18f0bbd47048a-Abstract.html},
  note      = {Alibaba Inc. Available at \url{https://github.com/alibaba/MNN}}
}

@misc{gerganov2023llama,
  author       = {Gerganov, Georgi},
  title        = {llama.cpp: {LLM} inference in {C/C++}},
  year         = {2023},
  month        = mar,
  publisher    = {GitHub},
  howpublished = {\url{https://github.com/ggerganov/llama.cpp}},
  url          = {https://github.com/ggerganov/llama.cpp},
  note         = {MIT License}
}

@misc{cmsis_nn,
  author       = {{Arm Limited}},
  title        = {CMSIS-NN: Efficient Neural Network Kernels for Arm Cortex-M CPUs},
  url          = {https://github.com},
  version      = {7.0.1},
  year         = {2025},
  date         = {2025-01-22},
  license      = {Apache-2.0},
  publisher    = {GitHub},
  repository   = {https://github.com/ARM-software/CMSIS-NN}
}

@misc{qualcomm_snpe,
  title        = {{Snapdragon Neural Processing Engine SDK}},
  author       = {{Qualcomm Technologies, Inc.}},
  organization = {Qualcomm Technologies, Inc.},
  year         = {2016},
  url          = {https://developer.qualcomm.com/software/qualcomm-neural-processing-sdk},
  note         = {Now branded as Qualcomm Neural Processing SDK for AI}
}

@misc{pytorch_mobile2019,
  author       = {{PyTorch Team}},
  title        = {{PyTorch Mobile}: End-to-End Workflow for Mobile Deployment},
  year         = {2019},
  month        = oct,
  organization = {Meta AI},
  url          = {https://pytorch.org/mobile/},
  note         = {Introduced in PyTorch 1.3. Now superseded by ExecuTorch}
}

@misc{metaai2024quantized,
  author = {{Meta AI}},
  title = {Introducing Quantized {Llama} Models with Increased Speed and a Reduced Memory Footprint},
  year = {2024},
  month = oct,
  day = {24},
  howpublished = {\url{https://ai.meta.com/blog/meta-llama-quantized-lightweight-models/}},
  note = {Accessed: 2025-10-29},
  organization = {Meta AI}
}

@misc{shao2024localglobalattentionadaptivemechanism,
      title={Local-Global Attention: An Adaptive Mechanism for Multi-Scale Feature Integration}, 
      author={Yifan Shao},
      year={2024},
      eprint={2411.09604},
      archivePrefix={arXiv},
      primaryClass={cs.CV},
      url={https://arxiv.org/abs/2411.09604}, 
}

@misc{kwon2023efficientmemorymanagementlarge,
      title={Efficient Memory Management for Large Language Model Serving with PagedAttention}, 
      author={Woosuk Kwon and Zhuohan Li and Siyuan Zhuang and Ying Sheng and Lianmin Zheng and Cody Hao Yu and Joseph E. Gonzalez and Hao Zhang and Ion Stoica},
      year={2023},
      eprint={2309.06180},
      archivePrefix={arXiv},
      primaryClass={cs.LG},
      url={https://arxiv.org/abs/2309.06180}, 
}

@misc{linux_foundation_annual_report_2024,
  author       = {The Linux Foundation},
  title        = {Annual Report 2024: Accelerating Industry Innovation},
  institution  = {The Linux Foundation},
  year         = {2024},
  url          = {https://www.linuxfoundation.org/resources/publications/linux-foundation-annual-report-2024},
  note         = {Accessed: 2025-10-30}
}

@misc{torchao,
  title={TorchAO: PyTorch-Native Training-to-Serving Model Optimization},
  author={torchao},
  url={https://github.com/pytorch/ao},
  license={BSD-3-Clause},
  month={oct},
  year={2024}
}

@misc{executorch_meta_foa_2025,
  title        = {{Accelerating On-Device ML on Meta's Family of Apps with ExecuTorch}},
  author       = {{Meta}},
  howpublished = {\url{https://engineering.fb.com/2025/07/28/android/executorch-on-device-ml-meta-family-of-apps/}},
  year         = {2025},
  note         = {Accessed: 2025-12-09}
}

@misc{executorch_rl_2025,
  title        = {{ExecuTorch Reality Labs On-Device AI}},
  author       = {{Meta}},
  howpublished = {\url{https://ai.meta.com/blog/executorch-reality-labs-on-device-ai/}},
  year         = {2025},
  note         = {Accessed: 2025-12-09}
}

@inproceedings{
wu2025control,
title={Control Flow Operators in PyTorch},
author={Yidi Wu and Thomas Ortner and Richard Zou and Edward Z. Yang and Adnan Akhundov and Horace He and Yanan Cao},
booktitle={Championing Open-source Development in ML Workshop @ ICML25},
year={2025},
url={https://openreview.net/forum?id=GMFG27v26J}
}
\bibliographystyle{mlsys2026}

\clearpage
\appendix

\section{LLM Operator-Level Performance Breakdown}
\label{sec:llm-breakdown}

To diagnose the performance gaps between ExecuTorch and llama.cpp observed in
Table~\ref{tab:llm_tps}, we profiled Llama~3.2~1B, Qwen3~0.6B, and Phi4~Mini on
the Samsung Galaxy S25 Ultra. ExecuTorch models use 8da4w quantization (int8
activations $\times$ int4 weights, group size 32); llama.cpp uses Q4\_0
(dynamically quantized int8 or fp activations $\times$ int4 weights, group size
32; 6-bit group-wise quantized weights for the LM head). All models are profiled
with a maximum context length of 2048, 256 prompt tokens, and 256 generated
tokens.

Llama~3.2~1B uses 16 layers with hidden dimension 2048 and 32 query heads,
resulting in a head dimension of 64; Qwen3~0.6B uses 28 layers, hidden
dimension 1024, 16 query heads, and an explicit head dimension of 128; and
Phi4~Mini uses 32 layers with hidden dimension 3072 and 24 query heads,
resulting in a head dimension of 128. All models use 8 KV heads.

Due to profiling overhead and run-to-run
variation, the profiling data presented in this Appendix may not line up exactly
with the measurements reported in Table~\ref{tab:llm_tps}. However, relative
performance between frameworks and backends should be fairly consistent.

In the tables below, columns associated with ``ggml'' represent inference with
llama.cpp.

\subsection{CPU: ExecuTorch XNNPACK vs llama.cpp CPU}

\begin{table}[t]
\centering
\small
\caption{CPU decode latency breakdown (ms/token) for 256 generated tokens.}
\label{tab:cpu_decode}
\resizebox{\columnwidth}{!}{%
\begin{tabular}{l rr rr rr}
\toprule
& \multicolumn{2}{c}{\textbf{Llama 3.2 1B}} & \multicolumn{2}{c}{\textbf{Qwen3 0.6B}} & \multicolumn{2}{c}{\textbf{Phi4 Mini}} \\
\cmidrule(lr){2-3} \cmidrule(lr){4-5} \cmidrule(lr){6-7}
\textbf{Category} & \textbf{ET} & \textbf{ggml} & \textbf{ET} & \textbf{ggml} & \textbf{ET} & \textbf{ggml} \\
\midrule
Linear              & 11.22 & 12.10 & 6.07  & 6.18  & 35.28 & 36.81 \\
Attention (SDPA)    & 1.95  & 1.71  & 5.37  & 2.39  & 10.01 & 4.36  \\
RMSNorm             & 0.07  & 0.13  & 0.12  & 0.22  & 0.25  & 0.46  \\
Activation (SwiGLU) & 0.12  & 0.12  & 0.04  & 0.08  & 0.85  & 0.31  \\
RoPE                & 0.00  & 0.03  & 0.00  & 0.06  & 0.00  & 0.10  \\
Other               & 0.58  & 0.11  & 0.51  & 0.26  & 1.56  & 0.34  \\
\midrule
\textbf{Total}      & \textbf{13.94} & \textbf{14.20} & \textbf{12.11} & \textbf{9.20} & \textbf{47.94} & \textbf{42.39} \\
\bottomrule
\end{tabular}%
}
\end{table}

\textbf{Decode} (Table~\ref{tab:cpu_decode}). Linear layers are at parity across
frameworks (11.22 vs 12.10~ms for Llama). The dominant gap is SDPA; llama.cpp's
implementation is 1.1--2.3$\times$ faster than XNNPACK's \texttt{custom\_sdpa}
for single-query decode (1.71 vs 1.95~ms on Llama, 2.39 vs 5.37~ms on Qwen, 4.36
vs 10.01~ms on Phi4). Note that for ET, many ops such as RMSNorm, Activation,
and RoPE are decomposed into core ATen ops, which makes it difficult to categorize/identify
operators produced by decompositions. As a result, RoPE is accounted for in
the ``Other'' category. The overhead introduced by decomposition also contributes
to a slight disadvantage for ExecuTorch.

\begin{table}[t]
\centering
\small
\caption{CPU prefill latency breakdown (total ms for 256 tokens).}
\label{tab:cpu_prefill}
\resizebox{\columnwidth}{!}{%
\begin{tabular}{l rr rr rr}
\toprule
& \multicolumn{2}{c}{\textbf{Llama 3.2 1B}} & \multicolumn{2}{c}{\textbf{Qwen3 0.6B}} & \multicolumn{2}{c}{\textbf{Phi4 Mini}} \\
\cmidrule(lr){2-3} \cmidrule(lr){4-5} \cmidrule(lr){6-7}
\textbf{Category} & \textbf{ET} & \textbf{ggml} & \textbf{ET} & \textbf{ggml} & \textbf{ET} & \textbf{ggml} \\
\midrule
Linear              & 295.58  & 314.39  & 138.88  & 165.59  & 1127.69 & 1311.05 \\
Attention (SDPA)    & 41.12   & 106.32  & 63.28   & 142.84  & 126.16  & 270.92  \\
RMSNorm             & 12.24   & 15.98   & 22.85   & 27.96   & 44.66   & 43.57   \\
Activation (SwiGLU) & 37.71   & 18.24   & 35.47   & 10.88   & 137.38  & 21.16   \\
RoPE                & 0.00    & 1.83    & 0.00    & 6.77    & 0.00    & 5.87    \\
Other               & 81.87   & 13.19   & 83.98   & 20.35   & 302.01  & 20.81   \\
\midrule
\textbf{Total}      & \textbf{468.52} & \textbf{469.95} & \textbf{344.46} & \textbf{374.39} & \textbf{1737.90} & \textbf{1673.38} \\
\bottomrule
\end{tabular}%
}
\end{table}

\textbf{Prefill} (Table~\ref{tab:cpu_prefill}). In contrast to decode, XNNPACK
is slightly faster in linear layers (6--19\% faster), which account for the
majority of inference time. Notably, for prefill ExecuTorch's
\texttt{custom\_sdpa} operator is 2.1--2.6$\times$ faster than
llama.cpp's attention implementation. Operator decompositions and inefficiencies
in some operator implementations such as embedding account for worse performance
in the ``Activation'' and ``Other'' categories. The net result of these effects is
that prefill performance between both frameworks is roughly on par.

ExecuTorch's \texttt{custom\_sdpa} operator uses a tiled
implementation to optimize for parallel processing of multiple queries. However,
minimal adjustments are made when handling single-token decode. In contrast,
llama.cpp's attention implementation makes significant
adjustments to its execution strategy when the batch dimension is 1. This may
explain why ExecuTorch's attention implementation is much faster for prefill,
but much slower for decode.

\subsection{GPU: ExecuTorch Vulkan vs llama.cpp OpenCL}

\begin{table}[t]
\centering
\small
\caption{GPU decode latency breakdown (ms/token) for 256 generated tokens. ``Dynamic Quant'' occurs only for
Vulkan, and consists of a ``choose quantization parameters'' shader and ``activation quantization'' shader
dispatched before each linear layer; for llama.cpp, ``Linear'' includes the
Q6\_K lm\_head kernel.}
\label{tab:gpu_decode}
\resizebox{\columnwidth}{!}{%
\begin{tabular}{l rr rr rr}
\toprule
& \multicolumn{2}{c}{\textbf{Llama 3.2 1B}} & \multicolumn{2}{c}{\textbf{Qwen3 0.6B}} & \multicolumn{2}{c}{\textbf{Phi4 Mini}} \\
\cmidrule(lr){2-3} \cmidrule(lr){4-5} \cmidrule(lr){6-7}
\textbf{Category} & \textbf{ET} & \textbf{ggml} & \textbf{ET} & \textbf{ggml} & \textbf{ET} & \textbf{ggml} \\
\midrule
Linear              & 13.04 & 19.80 & 7.03  & 7.57  & 40.78 & 71.18 \\
Dynamic Quant       & 0.63  & ---   & 0.78  & ---   & 1.30  & ---   \\
Attention (SDPA)    & 1.82  & 3.22  & 6.05  & 3.15  & 11.46 & 5.09  \\
RMSNorm             & 0.35  & 0.16  & 0.68  & 0.47  & 0.89  & 0.34  \\
Activation (SwiGLU) & 0.70  & 0.11  & 1.87  & 0.10  & 1.45  & 0.23  \\
RoPE                & 0.06  & 0.08  & 0.10  & 0.12  & 0.12  & 0.17  \\
Other               & 0.28  & 0.14  & 0.33  & 0.27  & 0.42  & 0.33  \\
\midrule
\textbf{Total}      & \textbf{16.88} & \textbf{23.51} & \textbf{16.84} & \textbf{11.67} & \textbf{56.43} & \textbf{77.35} \\
\bottomrule
\end{tabular}%
}
\end{table}

\textbf{Decode} (Table~\ref{tab:gpu_decode}): There are two notable factors that
explain differences in operator latency distribution between the two frameworks.
First, llama.cpp uses 6-bit group-wise quantized weights for the LM-head, and
the associated compute shader is 3.8--6.7$\times$ slower than ExecuTorch
Vulkan's corresponding 4-bit quantized linear compute shader (e.g.,\ 9.74 vs
2.58~ms for Llama, 40.86 vs 6.06~ms for Phi4). Note that due to the differences
in quantization, the computation being performed by both frameworks is not
the same.

Next, notice that while ExecuTorch Vulkan's SDPA implementation is faster in
Llama 3.2 1B (1.82 vs 3.22~ms), it is 1.9--2.3$\times$ slower in Qwen3 0.6B and
Phi4 Mini (6.05 vs 3.15~ms, 11.46 vs 5.09~ms). In the Vulkan backend, SDPA is
implemented by three shaders: one to compute the attention weight (i.e., matrix
multiply between query tensor and key cache fused with masking and scaling), one
to apply softmax to the attention weight, and one to compute the final matrix
multiplication between the attention weight and the value cache. The latency of
the first two shaders is fairly consistent across models, but the final shader
is much slower for Qwen3 and Phi4; furthermore, the latency increases
dramatically with context length (see Table~\ref{tab:sdpa_context}). Llama 3.2 uses a head dimension
of 64, while Qwen3 and Phi4 use a head dimension of 128, which doubles the
memory footprint of the value cache. For the value cache tensor, ExecuTorch
Vulkan keeps the head dim contiguous and the context dim (i.e., the reduction
dim) as the outermost dimension; this memory layout may be suboptimal and make
the implementation more susceptible to the increased memory pressure from the
doubled head dimension in Qwen3 and Phi4. Another significant factor is that llama.cpp uses
fp16 storage for cache tensors (compared to fp32 for ExecuTorch Vulkan), which helps alleviate the increased memory
pressure from larger cache tensors.

\begin{table}[t]
\centering
\small
\caption{Per-layer average SDPA latency ($\mu$s) at different context lengths for ExecuTorch Vulkan on the Samsung Galaxy S25.}
\label{tab:sdpa_context}
\resizebox{\columnwidth}{!}{%
\begin{tabular}{r l rrr}
\toprule
\textbf{Ctx} & \textbf{Sub-kernel} & \textbf{Llama 3.2 1B} & \textbf{Qwen3 0.6B} & \textbf{Phi4 Mini} \\
\midrule
300 & compute\_attn\_weights & 46.4 & 28.8 & 37.5 \\
300 & attn\_weights\_softmax & 6.5  & 5.9  & 6.1  \\
300 & compute\_out           & 35.6 & 100.8 & 143.5 \\
\cmidrule(lr){1-5}
400 & compute\_attn\_weights & 60.3 & 36.1 & 49.0 \\
400 & attn\_weights\_softmax & 7.2  & 6.2  & 6.7  \\
400 & compute\_out           & 45.0 & 182.9 & 232.4 \\
\cmidrule(lr){1-5}
500 & compute\_attn\_weights & 74.0 & 43.2 & 58.6 \\
500 & attn\_weights\_softmax & 7.3  & 6.3  & 6.6  \\
500 & compute\_out           & 62.1 & 259.2 & 404.2 \\
\bottomrule
\end{tabular}%
}
\end{table}

The net effect of the above two factors results in ExecuTorch Vulkan achieving
faster decode latency compared to llama.cpp, except for the Qwen3 0.6B
model where the increased SDPA latency outweighs the differences in linear
layers.

As with ExecuTorch XNNPACK, there are some operators (e.g., RMSNorm,
SiLU) that are currently decomposed by ExecuTorch when lowering to the Vulkan
delegate, which results in higher latencies for those categories when compared
to llama.cpp.

\begin{table}[t]
\centering
\small
\caption{GPU prefill latency breakdown (total ms, 256 tokens). ``Dynamic Quant'' occurs only for
Vulkan, and consists of a choose quantization parameters shader and activation quantization shader
dispatched before each linear layer.}
\label{tab:gpu_prefill}
\resizebox{\columnwidth}{!}{%
\begin{tabular}{l rr rr rr}
\toprule
& \multicolumn{2}{c}{\textbf{Llama 3.2 1B}} & \multicolumn{2}{c}{\textbf{Qwen3 0.6B}} & \multicolumn{2}{c}{\textbf{Phi4 Mini}} \\
\cmidrule(lr){2-3} \cmidrule(lr){4-5} \cmidrule(lr){6-7}
\textbf{Category} & \textbf{ET} & \textbf{ggml} & \textbf{ET} & \textbf{ggml} & \textbf{ET} & \textbf{ggml} \\
\midrule
Linear              & 212.26 & 205.13 & 98.30  & 104.26 & 828.40  & 669.29 \\
Dynamic Quant       & 11.42  & ---    & 10.31  & ---    & 32.94   & ---    \\
Attention (SDPA)    & 20.04  & 14.92  & 31.11  & 19.16  & 61.46   & 32.51  \\
RMSNorm             & 1.44   & 2.35   & 5.23   & 8.68   & 4.24    & 5.78   \\
Activation (SwiGLU) & 29.22  & 5.68   & 38.01  & 3.55   & 79.39   & 11.88  \\
RoPE                & 1.92   & 1.55   & 6.91   & 2.30   & 13.98   & 4.83   \\
Other               & 8.69   & 3.87   & 11.73  & 9.53   & 23.74   & 13.82  \\
\midrule
\textbf{Total}      & \textbf{284.99} & \textbf{233.50} & \textbf{201.60} & \textbf{147.48} & \textbf{1044.15} & \textbf{738.12} \\
\bottomrule
\end{tabular}%
}
\end{table}

\textbf{Prefill} (Table~\ref{tab:gpu_prefill}): llama.cpp has a clear advantage
in prefill driven by lower latency for linear layers and attention, which
account for the majority of inference time. llama.cpp's linear layers perform
fp16 accumulation with fp16 activations and dequantized 4-bit weights, while ExecuTorch
Vulkan dynamically quantizes fp32 activations to 8-bit and performs integer
accumulation using hardware-accelerated integer dot product instructions. Though
int8 compute throughput is theoretically double that of fp16 compute throughput,
Vulkan's linear layer shaders have longer latencies compared to llama.cpp, especially for Phi4 Mini. This is likely due to
the additional overhead of loading and applying per-group quantization
parameters during the requantization step; as seen in Table~\ref{tab:llm_tps},
using a group size of 128 compared to 32 results in a $\sim$25--30\% increase in
overall prefill throughput for Qwen3 and Llama 3.2, and a $\sim$56\% increase for
Phi4 Mini. The additional cost of dynamic quantization further compounds the
latency difference in linear layers between the two frameworks. For attention,
the same factors that contributed to worse latency in the decode step also apply
in prefill.

As with decode, operator decomposition accounts for more latency in Activation
operators. Additionally, the ``Other'' category for ExecuTorch Vulkan is dominated
by reshape operators, which wrap attention layers.

\clearpage
\section{Contributor Acknowledgements}
\label{sec:ack-appendix}
ExecuTorch is a collaborative effort spanning many teams and organizations. We
gratefully acknowledge the following contributors.

\textbf{Apple} \\[2pt]
Kulin Seth. \\[2pt]
Yifan Shen. \\[2pt]
Gyan Sinha. \\[2pt]
Denis Vieriu.

\textbf{Arm} \\[2pt]
Tom Allsop. \\[2pt]
Zingo Andersen. \\[2pt]
Oscar Andersson. \\[2pt]
Per Åstrand. \\[2pt]
Baris Demirbilek. \\[2pt]
Rob Elliott. \\[2pt]
George Gekov. \\[2pt]
Per Held. \\[2pt]
Agrima Khare. \\[2pt]
Benjamin Klimczak. \\[2pt]
Fredrik Knutsson. \\[2pt]
Emma Kujala. \\[2pt]
Sebastian Larsson. \\[2pt]
Xingguo Li. \\[2pt]
Martin Lindström. \\[2pt]
Adrian Lundell. \\[2pt]
Erik Lundell. \\[2pt]
Måns Nilsson. \\[2pt]
Michiel Olieslagers. \\[2pt]
Ryan O'Shea. \\[2pt]
Yufeng Shi. \\[2pt]
Saoirse Stewart. \\[2pt]
Charlie Stokes. \\[2pt]
Robert Taylor. \\[2pt]
Carey Williams. \\[2pt]
Elena Zhelezina.

\textbf{Cadence} \\[2pt]
Andrew Grebenisan. \\[2pt]
Chandana Madhira. \\[2pt]
The Cadence team.

\textbf{Intel} \\[2pt]
Aamir Nazir. \\[2pt]
Yamini Nimmagadda. \\[2pt]
Surya Siddharth Pemmaraju.

\textbf{MediaTek} \\[2pt]
The NeuroPilot team.

\textbf{NXP} \\[2pt]
Roman Janik. \\[2pt]
Robert Kalmar. \\[2pt]
Jiri Ocenasek. \\[2pt]
Martin Pavella. \\[2pt]
Davis Sawer. \\[2pt]
Šimon Strýček.

\textbf{Qualcomm} \\[2pt]
Felix Baum. \\[2pt]
Hao-Wei Hsu. \\[2pt]
Winston Kuo. \\[2pt]
Harsh Shah. \\[2pt]
Chun-I Tsai. \\[2pt]
Kiwi Wang. \\[2pt]
Sheng Feng Wu. \\[2pt]
YuYang Zhuang.

\textbf{Samsung} \\[2pt]
Collin Allen. \\[2pt]
Hoon Choi. \\[2pt]
Alex Dean. \\[2pt]
Mostafa El-Khamy. \\[2pt]
SangHyuck Ha. \\[2pt]
Fangming He. \\[2pt]
Shujie Huang. \\[2pt]
Bruce Kim. \\[2pt]
Sangsoo Ko. \\[2pt]
Pavan Lanka. \\[2pt]
Jiseong Oh. \\[2pt]
Sicheon Oh.

\textbf{Tencent} \\[2pt]
Jie Fu.

\textbf{Independent Contributors} \\[2pt]
Zuby Afzal. \\[2pt]
Xiang Li.

\textbf{Meta} \\[2pt]
In addition to the paper authors, we thank: \\[2pt]
Eli Amesefe. \\[2pt]
Stefano Cadario. \\[2pt]
Avik Chaudhuri. \\[2pt]
Xingying Cheng. \\[2pt]
Matthias Cremon. \\[2pt]
Salil Desai. \\[2pt]
Alban Desmaison. \\[2pt]
Huy Do. \\[2pt]
Riley Dulin. \\[2pt]
Soumyadeep Ghosh. \\[2pt]
Chris Gottbrath. \\[2pt]
Min Guo. \\[2pt]
Lunwen He. \\[2pt]
Nitin Jain. \\[2pt]
Svetlana Karslioglu. \\[2pt]
Harshit Khaitan. \\[2pt]
Ali Khosh. \\[2pt]
Ji Li. \\[2pt]
Yi Li. \\[2pt]
Juniper Pineda. \\[2pt]
Varun Puri. \\[2pt]
Jathu Satkunarajah. \\[2pt]
Nathanael See. \\[2pt]
Nikita Shulga. \\[2pt]
Jake Stevens. \\[2pt]
Michael Suo. \\[2pt]
Andrey Talman. \\[2pt]
Chris Thompson. \\[2pt]
Vivek Trivedi. \\[2pt]
Chakri Uddaraju. \\[2pt]
Eli Uriegas. \\[2pt]
Jesse White. \\[2pt]
Yidi Wu. \\[2pt]
Yiwen Xie. \\[2pt]
Justin Yip. \\[2pt]
Shangdi Yu.

We also thank these individuals for their contributions to ExecuTorch while working at Meta:

Ishan Aryendu. \\[2pt]
Michael Gschwind. \\[2pt]
Rohan Joshi. \\[2pt]
Akshit Khurana. \\[2pt]
Juntian Liu. \\[2pt]
Olivia Liu. \\[2pt]
Dhruv Matani. \\[2pt]
Bujji Setty. \\[2pt]
Joe Spisak. \\[2pt]
Conan Truong. \\[2pt]
Shen Chen Xu.


\end{document}